\documentclass[10pt,journal,compsoc]{IEEEtran}
\ifCLASSOPTIONcompsoc
  \usepackage[nocompress]{cite}

\usepackage{times}
\usepackage{epsfig}
\usepackage{graphicx}
\usepackage{amsmath}
\usepackage{amssymb}
\usepackage{multirow}
\usepackage{indentfirst}
\usepackage{mathrsfs}
\usepackage{caption}

\usepackage{algorithm}
\usepackage{algorithmic}
\else
  \usepackage{cite}
\fi
\ifCLASSINFOpdf
\else
\fi
\begin{document}
%
% paper title
% Titles are generally capitalized except for words such as a, an, and, as,
% at, but, by, for, in, nor, of, on, or, the, to and up, which are usually
% not capitalized unless they are the first or last word of the title.
% Linebreaks \\ can be used within to get better formatting as desired.
% Do not put math or special symbols in the title.
\title{A Hierarchical Distributed Processing \\ Framework for Big Image Data}
%
%
% author names and IEEE memberships
% note positions of commas and nonbreaking spaces ( ~ ) LaTeX will not break
% a structure at a ~ so this keeps an author's name from being broken across
% two lines.
% use \thanks{} to gain access to the first footnote area
% a separate \thanks must be used for each paragraph as LaTeX2e's \thanks
% was not built to handle multiple paragraphs
%
%
%\IEEEcompsocitemizethanks is a special \thanks that produces the bulleted
% lists the Computer Society journals use for "first footnote" author
% affiliations. Use \IEEEcompsocthanksitem which works much like \item
% for each affiliation group. When not in compsoc mode,
% \IEEEcompsocitemizethanks becomes like \thanks and
% \IEEEcompsocthanksitem becomes a line break with idention. This
% facilitates dual compilation, although admittedly the differences in the
% desired content of \author between the different types of papers makes a
% one-size-fits-all approach a daunting prospect. For instance, compsoc
% journal papers have the author affiliations above the "Manuscript
% received ..."  text while in non-compsoc journals this is reversed. Sigh.

\author{Le~Dong,~\IEEEmembership{Member,~IEEE,}
        ~Zhiyu~Lin,~Yan~Liang,~Ling~He,~Ning~Zhang,\\
        Qi Chen, Xiaochun Cao and~Ebroul~Izquierdo,~\IEEEmembership{Senior~Member,~IEEE}% <-this % stops a space
\IEEEcompsocitemizethanks{\IEEEcompsocthanksitem L. Dong, Z. Y. Lin, L. He, N. Zhang, are with the University of Electronic Science and Technology of China, Chengdu 611731, China.\protect\\
% note need leading \protect in front of \\ to get a newline within \thanks as
% \\ is fragile and will error, could use \hfil\break instead.
E-mail: ledong@uestc.edu.cn.
\IEEEcompsocthanksitem Y. Liang is with the China Academy of Engineering Physics, Mianyang, China.
\IEEEcompsocthanksitem Q. Chen is with the Cyberspace Administration of the People’s Republic of China, Beijing, China.
\IEEEcompsocthanksitem X. C. Cao is with the Institute of Information Engineering, Chinese Academy of Sciences, Beijing, China.
\IEEEcompsocthanksitem E. Izquierdo is with the Queen Mary, University of London, London E1 4NS, U.K.}% <-this % stops an unwanted space
\thanks{}}

% note the % following the last \IEEEmembership and also \thanks -
% these prevent an unwanted space from occurring between the last author name
% and the end of the author line. i.e., if you had this:
%
% \author{....lastname \thanks{...} \thanks{...} }
%                     ^------------^------------^----Do not want these spaces!
%
% a space would be appended to the last name and could cause every name on that
% line to be shifted left slightly. This is one of those "LaTeX things". For
% instance, "\textbf{A} \textbf{B}" will typeset as "A B" not "AB". To get
% "AB" then you have to do: "\textbf{A}\textbf{B}"
% \thanks is no different in this regard, so shield the last } of each \thanks
% that ends a line with a % and do not let a space in before the next \thanks.
% Spaces after \IEEEmembership other than the last one are OK (and needed) as
% you are supposed to have spaces between the names. For what it is worth,
% this is a minor point as most people would not even notice if the said evil
% space somehow managed to creep in.

% The paper headers
\markboth{Journal of \LaTeX\ Class Files,~Vol.~XX, No.~XX, XX~XXXX}%
{Shell \MakeLowercase{\textit{et al.}}: Bare Demo of IEEEtran.cls for Computer Society Journals}
% The only time the second header will appear is for the odd numbered pages
% after the title page when using the twoside option.
%
% *** Note that you probably will NOT want to include the author's ***
% *** name in the headers of peer review papers.                   ***
% You can use \ifCLASSOPTIONpeerreview for conditional compilation here if
% you desire.

% The publisher's ID mark at the bottom of the page is less important with
% Computer Society journal papers as those publications place the marks
% outside of the main text columns and, therefore, unlike regular IEEE
% journals, the available text space is not reduced by their presence.
% If you want to put a publisher's ID mark on the page you can do it like
% this:
%\IEEEpubid{0000--0000/00\$00.00~\copyright~2014 IEEE}
% or like this to get the Computer Society new two part style.
%\IEEEpubid{\makebox[\columnwidth]{\hfill 0000--0000/00/\$00.00~\copyright~2014 IEEE}%
%\hspace{\columnsep}\makebox[\columnwidth]{Published by the IEEE Computer Society\hfill}}
% Remember, if you use this you must call \IEEEpubidadjcol in the second
% column for its text to clear the IEEEpubid mark (Computer Society jorunal
% papers don't need this extra clearance.)

% use for special paper notices
%\IEEEspecialpapernotice{(Invited Paper)}

% for Computer Society papers, we must declare the abstract and index terms
% PRIOR to the title within the \IEEEtitleabstractindextext IEEEtran
% command as these need to go into the title area created by \maketitle.
% As a general rule, do not put math, special symbols or citations
% in the abstract or keywords.
\IEEEtitleabstractindextext{%
\begin{abstract}
This paper introduces an effective processing framework nominated ICP (Image Cloud Processing) 
to powerfully cope with the data explosion in image processing field. 
While most previous researches focus on optimizing the image processing 
algorithms to gain higher efficiency, our work dedicates to providing a 
general framework for those image processing algorithms, which can be 
implemented in parallel so as to achieve a boost in time efficiency without 
compromising the results performance along with the increasing image scale. 
The proposed ICP framework consists of two mechanisms, \emph{i.e.} SICP (Static ICP) 
and DICP (Dynamic ICP). Specifically, SICP is aimed at processing the big image data  
pre-stored in the distributed system, while DICP is proposed for dynamic input. 
To accomplish SICP, two novel data representations named P-Image and Big-Image are 
designed to cooperate with MapReduce to achieve more optimized configuration and higher efficiency.
 DICP is implemented through a parallel processing procedure working with the traditional 
 processing mechanism of the distributed system. Representative results of comprehensive 
 experiments on the challenging ImageNet dataset are selected to validate the capacity 
 of our proposed ICP framework over the traditional state-of-the-art methods, both in time efficiency and quality of results.
\end{abstract}

% Note that keywords are not normally used for peerreview papers.
}

% make the title area
\maketitle

% To allow for easy dual compilation without having to reenter the
% abstract/keywords data, the \IEEEtitleabstractindextext text will
% not be used in maketitle, but will appear (i.e., to be "transported")
% here as \IEEEdisplaynontitleabstractindextext when the compsoc
% or transmag modes are not selected <OR> if conference mode is selected
% - because all conference papers position the abstract like regular
% papers do.
\IEEEdisplaynontitleabstractindextext
% \IEEEdisplaynontitleabstractindextext has no effect when using
% compsoc or transmag under a non-conference mode.

% For peer review papers, you can put extra information on the cover
% page as needed:
% \ifCLASSOPTIONpeerreview
% \begin{center} \bfseries EDICS Category: 3-BBND \end{center}
% \fi
%
% For peerreview papers, this IEEEtran command inserts a page break and
% creates the second title. It will be ignored for other modes.
\IEEEpeerreviewmaketitle

\IEEEraisesectionheading{\section{Introduction}\label{sec:introduction}}
% Computer Society journal (but not conference!) papers do something unusual
% with the very first section heading (almost always called "Introduction").
% They place it ABOVE the main text! IEEEtran.cls does not automatically do
% this for you, but you can achieve this effect with the provided
% \IEEEraisesectionheading{} command. Note the need to keep any \label that
% is to refer to the section immediately after \section in the above as
% \IEEEraisesectionheading puts \section within a raised box.

% The very first letter is a 2 line initial drop letter followed
% by the rest of the first word in caps (small caps for compsoc).
%
% form to use if the first word consists of a single letter:
% \IEEEPARstart{A}{demo} file is ....
%
% form to use if you need the single drop letter followed by
% normal text (unknown if ever used by IEEE):
% \IEEEPARstart{A}{}demo file is ....
%
% Some journals put the first two words in caps:
% \IEEEPARstart{T}{his demo} file is ....
%
% Here we have the typical use of a "T" for an initial drop letter
% and "HIS" in caps to complete the first word.
\IEEEPARstart{O}{ver} recent years, image processing has gained wide attention due to its comprehensive applications in various areas, such as engineering, industrial manufacturing, military, and health, \begin{slshape}etc..\end{slshape} However, in spite of its expansive development prospect, huge data amount comes along and hence triggers severe constraints on data storage and processing efficiency, which calls for urgent solution to relieve such limitations. Particularly, since Web age and search engine started to develop and boom, most real business Web sites like Google, Baidu, Twitter, Facebook, \begin{slshape}etc.\end{slshape} have to deal with millions of users' requests for image storage, indexing, querying and searching within acceptable time. Furthermore, the prosperity of big image data over recent years has undoubtedly aggravated the challenge that current image processing field commonly faces. To this end, arduous efforts from related research fields have been made so far to propose high-efficiency image processing algorithms. Nonetheless, most of these efforts are only focused on optimizing the image processing algorithms, while totally neglecting the inherent deficiency of the single node based processing procedure. Therefore, although previous works [1-9] do have made some advancements to release the difficulties that the image processing field faces, their performance is commonly limited to a rather low level due to the inefficient processing based on a single machine.
% You must have at least 2 lines in the paragraph with the drop letter
% (should never be an issue)
What deserves to be noticed is that huge quantities of image data are typically stored in a distributed system accompanied by the growing popularity of big data. On the other hand, cloud computing is a prevalent commercial infrastructure paradigm that promises to eliminate the need for maintaining expensive computing facilities by companies and institutes alike. [10] has analyzed the performance of cloud computing services for scientific computing workloads and drew the conclusion that while current cloud computing services are insufficient for scientific computing at large, they may still be a good solution for the scientists who need resources instantly and temporarily. It is therefore significant to take full advantages of the abundant computing resources that the distributed system offers in a cloud processing manner. Given the distributed resources, parallel processing can undoubtedly achieve state-of-the-art improvements when compared with the traditional processing methods limited to a single machine, yet meanwhile, this attempt is a demanding challenge.
In recent years, considering the high efficiency that parallel processing brings, researchers have attempted to propose image processing algorithms that can be implemented in parallel, among which image classification [11], feature extraction and matching [11-12] can serve as representative instances. All those algorithms can run on multiple nodes in parallel and hence significantly improves the time efficiency. At present, however, few general frameworks are available for such image processing algorithms that should have obtained better performance for their parallelism. In 2010, Facebook proposed a framework called Haystack that has accomplished significant improvements in huge image data storage. However, since that the real applications call for both storage and efficient processing, Haystack just cannot provide desirable solutions for its lacking of the ability to process big image data effectively. In this sense, a general framework aimed at both big image data storage and effective processing is highly demanded. To this end, MapReduce[14] deserves to be mentioned.

MapReduce[14] is recognized as a popular framework to handle huge data amount in the cloud environment due to its excellent scalability and fault tolerance. Application programs based on MapReduce can work on a huge cluster of thousands of desktops and reliably process Peta-Bytes data in parallel. Owing to this, time efficiency successfully gains a desirable improvement. Until now, MapReduce has been widely applied into numerous applications including data anonymization [15], text tokenization, indexing and searching, data mining [16], machine learning [17], \begin{slshape}etc..\end{slshape} Aimed at achieving higher time efficiency on the associated applications, recent endeavors involve many industry giants to make efforts by leveraging MapReduce. For example, Yahoo has been working on a couple of real-time analytic projects, including S4 and MapReduce Online [18]. In addition, IBM has been devoted to developing real-time products such as InfoSphere Streams [20] and Jonas’s Entity Analytics software [21] used to analyze stream data more accurately. Despite that these frameworks have successfully implemented the efficient processing of text data and stream data, they do little contribution to image processing field. Motivated by these cases' achievements and restrictions, we aim to implement an effective processing framework for big image data by the utilization of the cloud computing ability that MapReduce provides.

In this paper, we present and analyze a novel effective distributed framework named ICP (Image Cloud Processing) which is dedicated to offering a reliable and efficient model for vision tasks [32-35]. The core design of ICP is to utilize the affluent computing resources provided by the distributed system so as to implement effective parallel processing. The elegant distributed processing mechanism that ICP contains is defined from two comprehensive perspectives: 1) efficiently processing those static big image data already stored in the distributed system, such as the task of image classification, image retrieval, \begin{slshape}etc.\end{slshape} that do not demand immediate response to the users but an efficient processing instead;  2) timely processing those dynamic input which needs to be processed immediately and return an immediate response to the users, especially for the requests from the mobile terminal, \emph{e.g.}, the image processing software in the users's mobile phone. Correspondingly, we call these two processing mechanisms SICP and DICP, where S and D denotes Static and Dynamic, respectively.

In order to accomplish the superiority of SICP which focuses on employing the distributed resources to achieve cloud computing, we propose two novel image data representations named P-Image and Big-Image which realize their potential with the joining efforts of MapReduce. P-Image, where P denotes Pure, only contains the necessary information including the filename, the pixel values, and the width-height of the input image, all of which are gained by decoding the initial input. Big-Image is a special representation of file which is large enough to contain a data file and an index file employed to store the P-Images and record their corresponding offsets, respectively. With the effective indexing structure, we can locate the P-Images at a high speed and hence improve the time efficiency of the whole processing procedure. In our SICP mechanism, Big-Image will replace the role of traditional small image files to act as input. Concretely, Big-Image will be partitioned into several groups to be processed in parallel by utilizing the sufficient computing resources offered by the distributed system. From this perspective, the design of Big-Image will contribute a lot to boost time efficiency without compromising the load performance. As for the DICP mechanism, we design a Master Proxy and a Matching Module to achieve effective processing by making these two cooperators to work with the inherent processing procedure of the distributed system in parallel. Briefly, the Master Proxy receives requests from the users and transmits the refined accessible parameters (\emph{e.g.} image filename and file extension) to the Matching Module. Then, the configuration file in the Matching Module will be used to match these parameters and determines whether the scheduling mechanism should call the related response algorithms. Thus, high stability and pressure resistance can be obtained via the DICP mechanism when processing dynamic input.

In general, the significant contributions of this paper can be summarized as follows:

\textbf{1) The proposed ICP is implemented in parallel and provides a general framework for image processing and achieves a boost in time efficiency without compromising the performance;}

\textbf{2) SCIP is aimed at efficiently processing large-scale images that have already been stored in the distributed system. Especially, two novel image processing algorithms named  P-images and Big-images are designed to avoid the repeated and time-consuming decoding operation, as well as to release memory consuming;}

\textbf{3) A complementary mechanism named DICP is applied for the new-coming image files. The stability and pressure resistance of DICP enable the requirement of dynamic input and urgent processing.}

The rest of this paper is organized as follows: Section 2 highlights the related work. Section 3 presents an overview of the ICP framework proposed in this paper. The SICP mechanism is elaborated in Section 4. Section 5 details the DICP. Experimental evidence that validates our work outperforms other techniques is presented in Section 6. Finally, Section 7 concludes the paper with directions for future work.

\begin{figure*}[ht]
\begin{center}
%\fbox{
\scalebox{1.1}[1.1]{\includegraphics[width=0.8\linewidth]{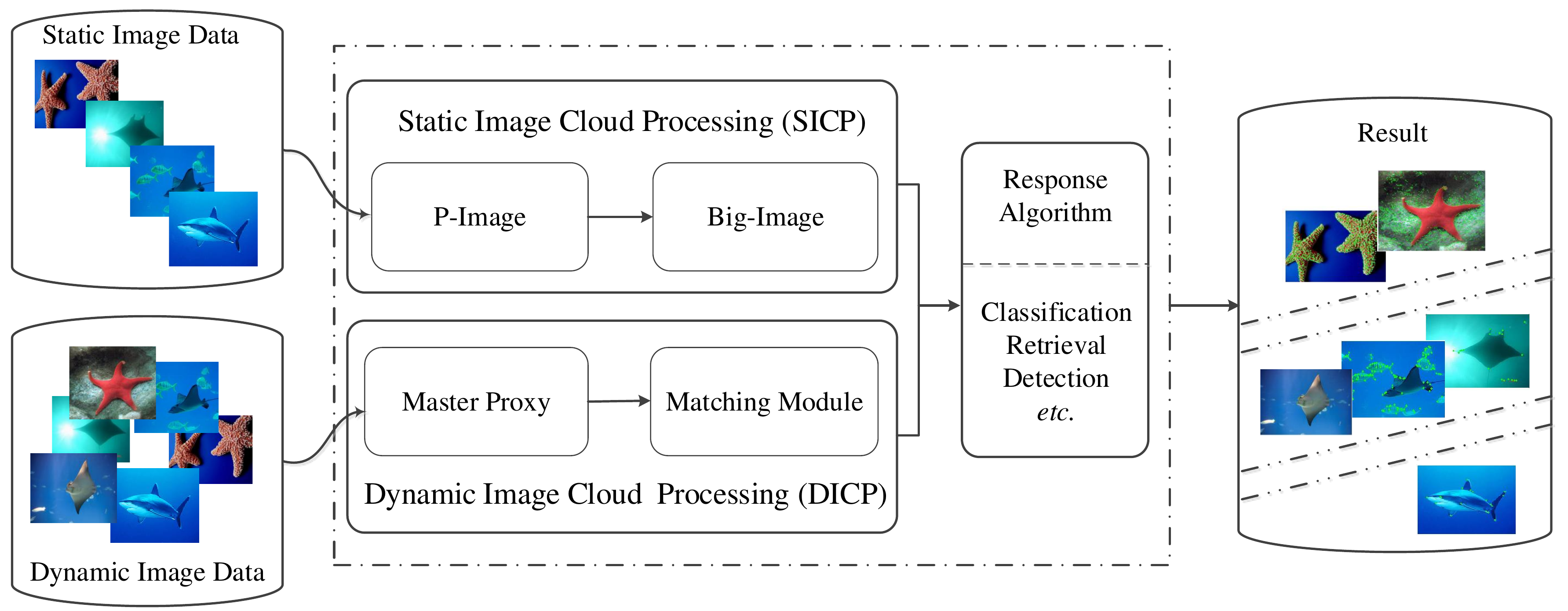}}
   %%The overview of our large-scale image retrieval system.
\end{center}
   \caption{ICP framework. Static image data is stored in the distributed system, followed by gaining the image data representation (P-Image and Big-Image), response algorithms can be called to process the static image data. Dynamic image data is mainly distributed in users' mobile terminals, followed by a Master Proxy and Matching Module, related response algorithms will be called according to the input parameters.}
\label{fig.long} \label{fig.onecol}
\end{figure*}

%\hfill mds

%\hfill September 17, 2014
\section{Related Work}
No doubt that big data has successfully become a central theme applied to large-scale computing problems in recent years. Researches based on parallel computing and distributed system have been carried out gradually, among which BigTable [22] and GFS [23] proposed by Google are typical achievements. Another representative work, \emph{i.e.}
MapReduce [14], is capable of processing huge data amount in a parallel distributed manner across numerous nodes. Just as well introduced in [16][19], there are three main processing phases in MapReduce: the Map phase, the Shuffle phase, and the Reduce phase. During the Map phase, the input data is distributed across the mapper machines, where each machine then processes a subset of the data in parallel and produces some $<key, value>$ pairs for each data record. Then, during the Shuffle phase, these gained $<key, value>$ pairs are repartitioned and sorted within each partition so that values corresponding to the same key can be grouped together into a values set $\left\{ v1, v2, \dots\right\}$. Finally, during the Reduce phase, each reducer machine processes a subset of the $<key, \left\{v1, v2, \dots \right\}>$ in parallel and writes the final results to the distributed file system.

As the most widely known open-source architecture of MapReduce, Hadoop [18][24] provides a reliable platform for researches demanding high efficiency, huge storage, and accurate analysis. By making use of Hadoop, significant improvements have been achieved in many aspects, such as the high time efficiency of file accessing and the requirement for real-time processing [25]. Besides, Hadoop has also been adopted to prompt the development of large-scale image processing. At present, attempts combined with Hadoop to implement big image data processing mainly include two alternative approaches. The first one is to regard Hadoop distributed computing framework as an effective tool to reduce the time consumption of image processing.

For instance, [26] utilizes Hadoop to extract the SIFT [13] (Scale-invariant feature transform) features and produce inverted index files [27]; [11] employs Hadoop to achieve image feature extraction and SVM training. Although both of these methods have effectively improved the time efficiency of image processing by the simple use of Hadoop, they lack adequate performance in processing big image data due to their ignorance of the fact that Hadoop is mainly developed for massive text processing. Without any additional designs, it is hard to show the advantages of Hadoop in processing large-scale images. Another alternative solution is devoted to enhancing the ability of Hadoop in processing images, namely, up to the text processing. The key point of this method is to convert the image data to binary data stream at first and then process these image data using the built-in data type of Hadoop (\emph{e.g.} BinaryWritable).

Note that image processing algorithms based on this method often need to employ the surrounding pixel points around the central one, whereas the traditional serialized processing of image data does not support this operation.Some researchers have successfully improved the performance of Hadoop by implementing the customized image data interface, recent examples of such method include [28-29]. These approaches have showed their progress in making the related image I/O formats distinguished by Hadoop, however, distinct algorithms are required to implement the conversion among different formats of image data. Furthermore, these methods neglect the parallelism and efficiency of the image processing algorithms based on Hadoop platform.

Inspired by the deficiency and challenges of the existing works, we propose a novel distributed processing framework named ICP to achieve high time efficiency both in accessing and processing the big image data stored on the cloud. While drawing on ideas from aforementioned efforts, our goal is quite different because we seek not only huge image scale but also the time efficiency, which leads to very different designs. Section 3 to Section 5 will elaborate the elegant design of our ICP.
\section{System Overview}

Our ICP framework consists of two complementary processing mechanisms, \emph{i.e.}, SICP (Static Image Cloud Processing) and DICP (Dynamic Image Cloud Processing). As shown in Fig.1, SICP is aimed at processing those large-scale image data that have been stored in the distributed system. Decode these static images first to maintain the necessary information as their corresponding P-Images which will be then stored in the data file contained in Big-Image. Then, when image processing is required, we just need to index the index file also stored in Big-Image to find the demanded P-Images which provide the needed image information. Given the needed image information, we can then implement the related image processing algorithms aimed at image classification, retrieval, detection, \begin{slshape}etc..\end{slshape} As for DICP, it is designed for the dynamic requests from the clients and must be able to return the results immediately. Master Proxy accepts the client's processing request and delivers it to the Master working in the inherent mechanism that the traditional distributed system owns. In parallel, Master Proxy transmits the accessible parameters (\emph{e.g.} the image filename and file extension) refined from the requests to the Matching Module in which these parameters will be matched with that set beforehand according to real applications. If successfully matched, the related response algorithms would be called and use the information provided by the inherent Master-Slave mechanism of the conventional distributed system to accomplish corresponding image processing operation. From the working mechanism of SICP and DICP, it is clear that SICP is fit for processing the big image data pre-strored in the distributed system and real time is not seriously demanded, while DICP is more applicable when millions of mobile terminals simultaneously make a request of image processing and demand for immediate response. Details of SICP and DICP will be elaborated in subsequent sections.

\section{Static Image Cloud Processing}

Just as aforementioned, SICP is an effective distributed processing mechanism dedicated to processing the big image data that have already been stored in the distributed system. In some sense, SICP may contribute more to those real business Web sites like Facebook, Google, Baidu, \begin{slshape}etc.\end{slshape} which impose a high demanding on efficiently processing large-scale image data. In order to take full advantages of SICP, MapReduce is employed to cooperate with the newly designed P-Image and Big-Image to implement large-scale and parallel processing in the cloud computing manner. Note that our SICP is not limited to implement on MapReduce, any other parallel processing framework is available, while MapReduce may provide the relatively highest superiority due to its excellent fault tolerance and load balancing. Meanwhile, industry giants like Google \begin{slshape}etc..\end{slshape} have proved the high scalability of MapReduce in real applications, likewise, our SICP can be surely scalable to a more complex environment.

\subsection{Modeling for P-Image and Big-Image}

Traditional image processing methods based on a single node need to decode the images and store all of the gained image information in memory. From this perspective, the image scale would be seriously restricted to a low level due to the limited memory space. Besides, when the processing is completed, the image information stored in memory will be lost and thus, it would demand another decoding when the image information is required again. Such repeated decoding operations would undoubtedly drag down the time efficiency of the whole processing procedure. In addition, storing uncompressed big image data in the distributed system will incur data redundancy. Here, we design P-Image and Big-Image to release these constraints. Just as depicted in Fig. 2, P-Image is actually the compressed version of the original image, which only contains the necessary image information
obtained by decoding the initial input. The information that reserved in P-Image includes the filename, the pixel values, and the width-height of the initial image. To our best knowledge, almost all of the image processing algorithms in computer vision are based on pixel information. Thus, the information contained in P-Image is enough for most of the image processing requirements. Once contained in P-Image, these information would not get lost and hence, time consumption will be greatly
reduced by avoiding the repeated decoding operations. In our design, we utilize a
\begin{figure}[h]
\begin{center}
%\fbox{
\scalebox{0.5}[0.5]{\includegraphics[width=0.8\linewidth]{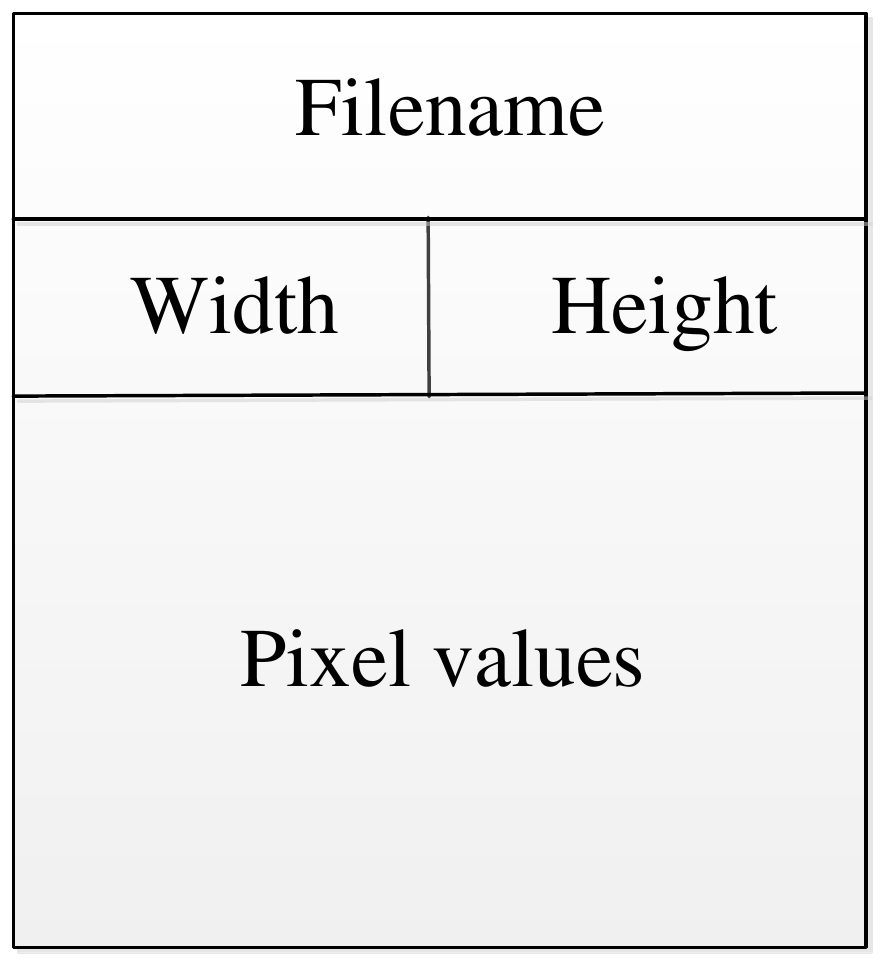}}
   %%The overview of our large-scale image retrieval system.
\end{center}
   \caption{The structure of P-Image.} \label{fig:long}
\label{fig:onecol}
\end{figure}
two-dimensional matrix to store the pixel values corresponding to those stored in
the P-Image. By accessing the matrix, we can obtain the pixel values at a high speed
owing to the one-to-one correspondence between the pixel coordinates recorded in the matrix and those contained in P-Image.

\begin{figure}[h]
\begin{center}
%\fbox{
\scalebox{0.9}[0.9]{\includegraphics[width=0.8\linewidth]{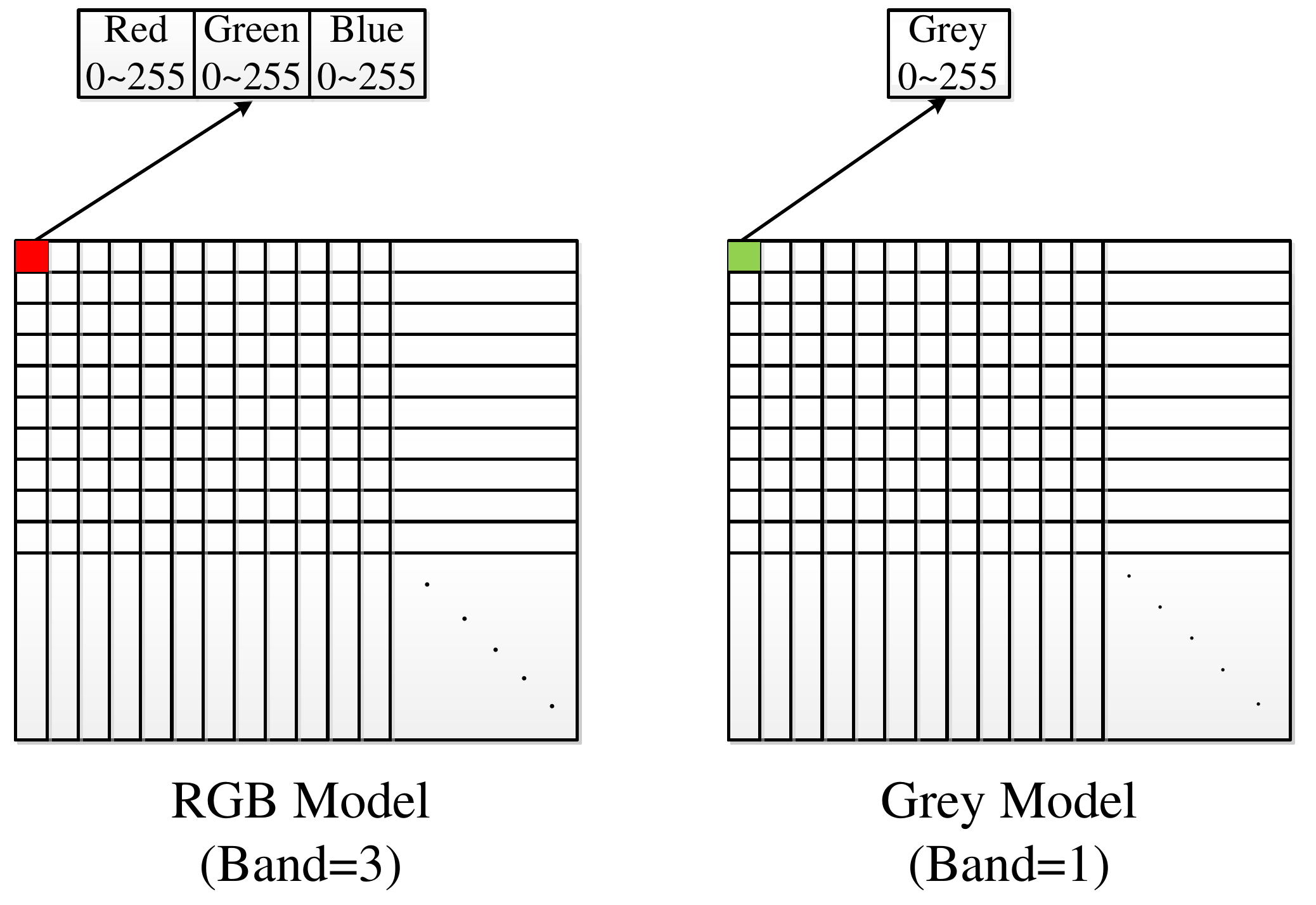}}
   %%The overview of our large-scale image retrieval system.
\end{center}
   \caption{The differences between RGB color mode and Grey color mode.}
\label{fig:long} \label{fig:onecol}
\end{figure}

In image processing field, RGB and Grey color modes are the most widely used color space to represent images. As depicted in Fig. 3, RGB color mode contains a $1 \times 3$ array to store the key of the three channels. By contrast, Grey color mode only contains a single key. Despite this difference, the key ranges from 0 to 255 no matter what the color mode is. Sometimes, we need to transform the RGB mode to Grey mode when using P-Image. In our work, we employ a famous formula of psychology to accomplish the transformation:
\begin{equation}
\begin{split}
M(x,y)=&M(x,y)_R\times 0.2989+M(x,y)_G\times 0.5870+ \\
                             &M(x,y)_B\times 0.1140,
\end{split}
\end{equation}
where \begin{math}M(x, y)\end{math} is a two dimensional matrix in which the elements are Grey pixel values, \begin{math}M(x, y)_R, M (x, y)_G, M (x, y)_B\end{math} represents the value of red channel, the value of green channel and the value of blue channel, respectively. In our design, different key is stored in distinct color mode.

\begin{figure}[h]
\begin{center}
%\fbox{
\scalebox{1.0}[1.0]{\includegraphics[width=0.8\linewidth]{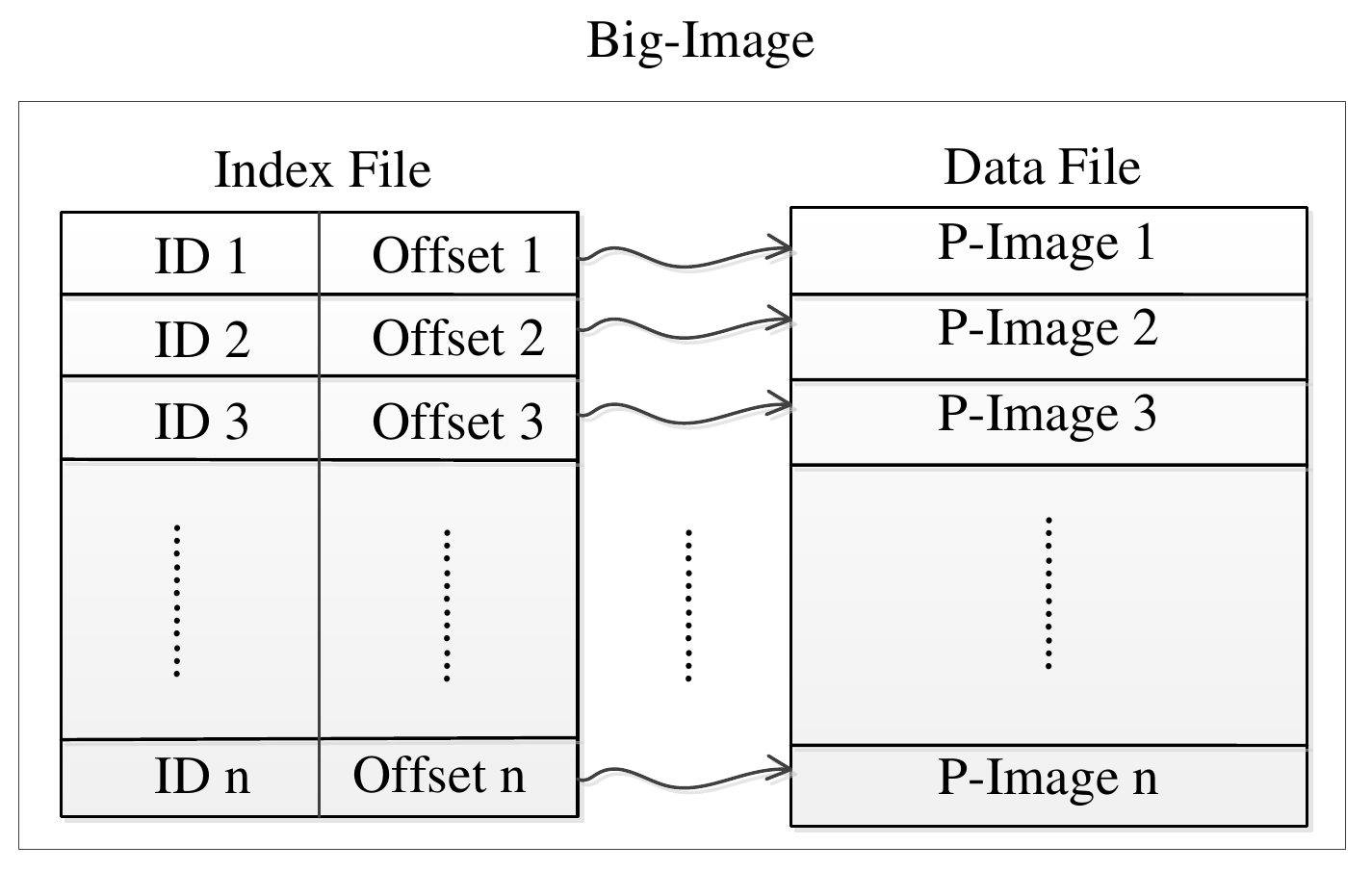}}
   %%The overview of our large-scale image retrieval system.
\end{center}
   \caption{The structure of Big-Image.}
\label{fig:long} \label{fig:onecol}
\end{figure}

Fig. 4 presents the structure of Big-Image which consists of a data file and an index file. The data file is employed to store the aforementioned P-Images, and the index file is utilized to record the ID and Offset of each P-Image stored in the data file. Here, we store the P-Images in Big-Image so as to save memory space, avoid a loss of image information, and process huge amount of images at a time. The catalogue of the index file is made up of two fields, \emph{i.e.} ID and Offset. The P-Image ID is computed by the Hash function with the P-Image filename, and the P-Image Offset denotes its corresponding location in the data file. Indexing through the index file using the ID to get the corresponding Offset, we can directly get the P-Images stored in the data file to extract the needed image information for subsequent processing. Compared with
the traditional small image files, Big-Image effectively avoid
the queueing delay. Users can set a threshold controlling the size of Big-Image according to the real applications. If the size of Big-Image is bigger than the threshold, then a new level of file can be designed to store multiple Big-Image files. The index structure is similar to that of Big-Image indexing P-Image.

\begin{figure}[h]
\begin{center}
%\fbox{
\scalebox{1.3}[1.4]{\includegraphics[width=0.8\linewidth]{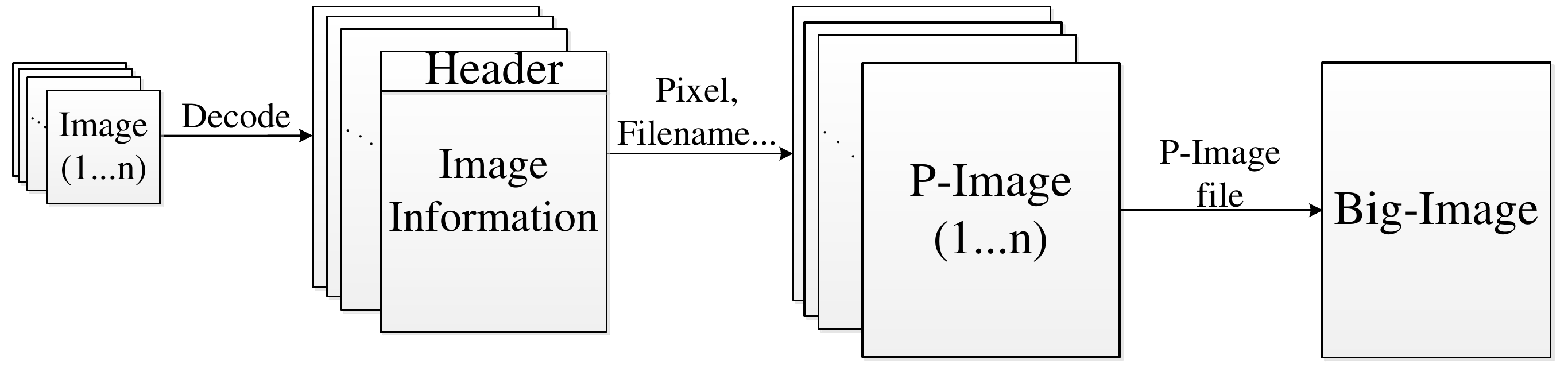}}
   %%The overview of our large-scale image retrieval system.
\end{center}
   \caption{Image data flow of P-Image and Big-Image.} \label{fig:long} \label{fig:onecol}
\end{figure}

Fig. 5 illustrates the image data flow of P-Images and Big-Image. Specifically, we obtain the P-Images by decoding the input images at first to reserve the necessary information including the filename, the width-height and the pixel values, and then store these P-Images into Big-Image waiting to be processed in subsequent stages. Note that producing P-Image and updating Big-Image can be implemented in parallel with the  image processing already underway. Concretely, when new image data comes, SICP decodes these images to produce P-Images and update Big-Image, while at the same time, the already started image processing can keep going on. Furthermore, Big-Image replaces the roles of traditional small image files acting as input, the efficiency of which will be validated in Section 6.3.

\subsection{Big Image Data Processing on SICP}

Traditional image processing methods usually utilize the small image files as serial processing units, which seriously limits the processing efficiency and even results in a breakdown once the cluster fails to timely process such huge amounts of small image files. To overcome these constraints, our SICP mechanism provides a different design. Algorithm 1 gives the pseudo-code for the implementation of SICP and Fig.6 illustrates the big image data processing on SICP.

\begin{algorithm}
\caption{SICP Mechanism }
\label{alg1}

\begin{center}\textbf{~~/* \emph{Step 1: Produce P-Images} */}\end{center}

~~~~~P-Image: an image representation containing filename,

~~~~~width-height, and pixel values;

~~~~~Input: Image[$1\dots$N];
\begin{algorithmic}[1]
%\STATE P-Image: an image representation containing filename, width-height, and pixel values
%\STATE Input: Image[1\dots N]
\FOR{\textbf{each} Image[$i$]}
  \STATE  Decode Image[$i$] to retain its filename, width, height, and pixel values;
  \ENDFOR \vspace{3ex}
\begin{center}\textbf{/* \emph{Step 2: Produce Big-Image} */}\end{center}

Big-Image: a huge file containing a data file and an index file;

ID: image filename; Offset: P-Image[i]'s size;

Offset$\leftarrow$0;

%\algstore{bkbreak}
\FOR{\textbf{each} P-Image[$i$]}
  \STATE  Offset$\leftarrow$Offset$+$the size of P-Image[$i$];
  \STATE  Insert ID and Offset of P-Image[$i$] into Big-Image.index;
  \STATE  Insert P-Image[$i$] into Big-Image.data;
  \ENDFOR \vspace{2ex}

%  \algstore{bkbreak}

\begin{center}\textbf{/* \emph{Step 3: Partition Big-Image} */}\end{center}

GP[$k$]: the $k^{th}$ group of P-Images;

BLOCKSIZE: the size of GP$[k]$;

NumMapTask: the number of Map Nodes;

NumMapTask$\leftarrow$Big-Image.size$/$BLOCKSIZE$+$$1$;

Offset$\leftarrow0$; count$\leftarrow1$; $i\leftarrow0$; $k\leftarrow1$;
%\algstore{bkbreak}
\WHILE{NumMapTask$>0$}
\WHILE{Offset$<$BLOCKSIZE$*$count}
\STATE Extract the P-Image[$i$] from Big-Image.data according to P-Image[$i$] Offset;
\STATE Insert P-Image[$i$] into GP[$k$];
\STATE Offset$\leftarrow$Offset$+$size of P-Image[$i$];
\STATE $i\leftarrow i+1;$
\ENDWHILE
\STATE $k\leftarrow k+1;$
\STATE Allocate GP[$k$] to the corresponding Map Node;
\STATE NumMapTask$\leftarrow$NumMapTask$-1$; count$\leftarrow$count$+1$;
\ENDWHILE
\end{algorithmic}
\end{algorithm}

\textbf{Step 1: Produce P-Images.} Since that all of the images are  typically represented by the structured data after encoding, 
we firstly produce P-Images. Different from the traditional image processing procedure, P-Image, which is composed of the extracted necessary information, 
successfully helps to avoid the repeated and time-consuming decoding operation and most importantly, it helps to release memory demanding by 
storing the numerous P-Images in the hard disk.

\textbf{Step 2: Produce Big-Image.} After Step 1, we design a special representation of file called Big-Image to store all of the gained P-Images. Big-Image consists of a data file to store the P-Images and an index file to store the corresponding ID and Offset (Line 4 to 8). Owing to the elegant design of the index structure, Big-Image contributes a lot to rapidly locate the P-Images required for processing. Besides, Big-Image contributes a lot to reduce the disk I/O when compared with conventional small files.

\textbf{Step 3: Partition Big-Image.} The core of our SICP lies in the parallel processing on a cluster of machines by utilizing the computing resources provided by the distributed system. Therefore, the single Big-Image needs to be partitioned into several groups to be processed on the Map Nodes (see the introduction of MapReduce in Section 2) in parallel. The P-Image amount in each group can be set in accordance with the real applications. Search the P-Images stored in the data file via their corresponding Offsets, and then insert these P-Images into one group (Line 10 to 15).

\begin{figure*}[ht]
\begin{center}
%\fbox{
\scalebox{1.3}[1.4]{\includegraphics[width=0.8\linewidth]{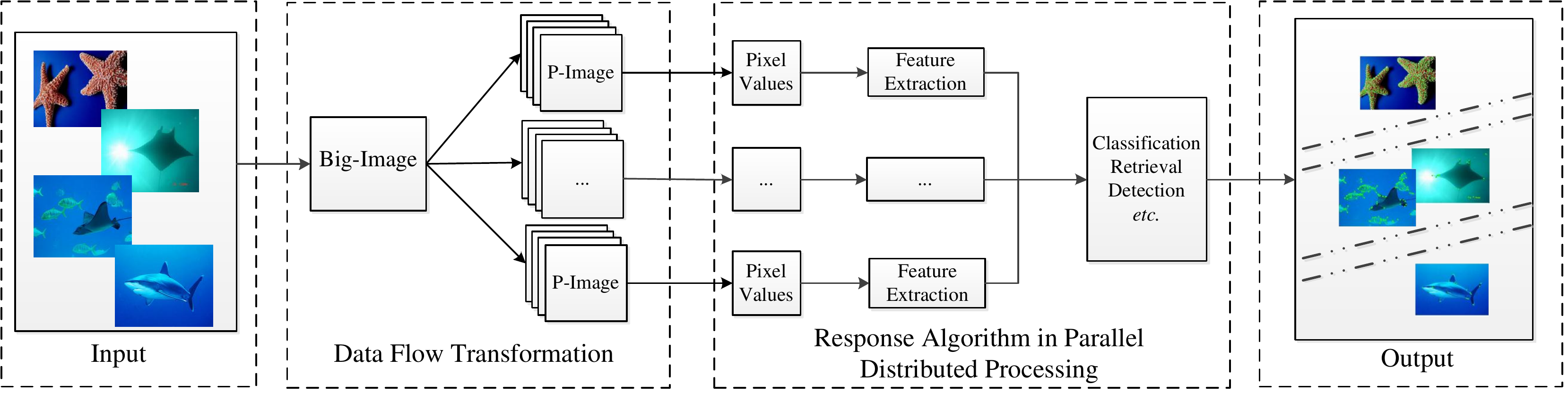}}
   %%The overview of our large-scale image retrieval system.
\end{center}
   \caption{The SICP mechanism.} \label{fig:long} \label{fig:onecol}
\end{figure*}
After the partition of Big-Image, each Map Node would deal with its corresponding \begin{math}GP[k]\end{math} (the \begin{math}k^{th}\end{math} group of P-Images) in parallel to gain pixel values and accomplish feature extraction. We call this processing procedure mapping function, \emph{i.e.} \begin{math}\mathscr{M} \left\{\cdot\right\}\end{math}, which takes \begin{math}GP[k]\end{math} as input. The eventually gained features can be defined as

\begin{equation}
{F_I}(k) = \mathscr{M}\left\{ {GP[k]} \right\},
\end{equation}
where \begin{math}F_I(k)\end{math} represents the total features of each \begin{math}GP[k]\end{math} after the \begin{math}\mathscr{M} \left\{\cdot\right\}\end{math} operation. The gained \begin{math}F_I(k)\end{math} would act as input to the reducing function, \emph{i.e.} \begin{math}\mathscr{R} \left\{\cdot\right\}\end{math}, in which \begin{math}\alpha_k\end{math} is another input coefficient(we will define the role of \begin{math}\alpha_k\end{math} shortly). The final output \begin{math}\mathscr{O}\end{math} is expressed as equation (3),

\begin{equation}
\mathscr{O} = \mathscr{R}\left\{ \sum_{k=1}^{NumMapTask} \alpha_k F_I(k)\right\}.
\end{equation}

 In the reducing operation, the features $F_I(k)$ obtained from each \begin{math}\mathscr{M} \left\{\cdot\right\}\end{math} would be collected for the subsequent processing, such as classification, retrieval, detection, \begin{slshape}etc..\end{slshape} In equation (3), \begin{math}\alpha_k\end{math} can be set as 1 or 0 according to the specific response algorithms. For instance, image classification and image retrieval require all the \begin{math}F_I(k)\end{math} and therefore, each \begin{math}\alpha_k\end{math} would be set as 1; for image detection which only needs one \begin{math}F_I(k)\end{math}, only the related \begin{math}\alpha_k\end{math} would be set as 1 while all of the others would be set as 0.

 According to the well-designed working mechanism of SICP elaborated above, the boost in time efficiency of our work largely results from the parallel processing. All of the operations, \emph{i.e.}, partitioning Big-Image, gaining pixel values, extracting features, mapping, reducing, \begin{slshape}etc.\end{slshape}, can be implemented in parallel. Apparently, compared with traditional methods taking single image file as input to be processed serially, P-Image and Big-Image contribute a lot to the parallel processing
in SICP mechanism. Instead of suffering a breakdown of the cluster, SICP guarantees a stable processing procedure even when the image scale reaches a huge extent, which largely attributes to the high scalability that the cloud computing platform owns. With the excellent cloud computing capability, the image processing equals to processing several big image files after the simple partition of Big-Image. In Section 6.4, convincing experimental results would powerfully validate the high efficiency of SICP.

\subsection{Dynamic Image Cloud Processing}

Since that SICP is proposed mainly for efficiently processing large-scale images that have already been stored in the distributed system, then, we wonder what if the new-coming image files require an urgent processing and immediate response. To solve this problem, we propose a complementary processing mechanism named DICP to perfectly meet the requirement of dynamic input and urgent processing. Different from SICP, DICP may mainly benefit the small requests from the mobile terminals, while the image data amount still keeps huge when considering the millions of simultaneous terminal requests.

\begin{figure}[h]
\begin{center}
%\fbox{
\scalebox{1.0}[1.0]{\includegraphics[width=0.8\linewidth]{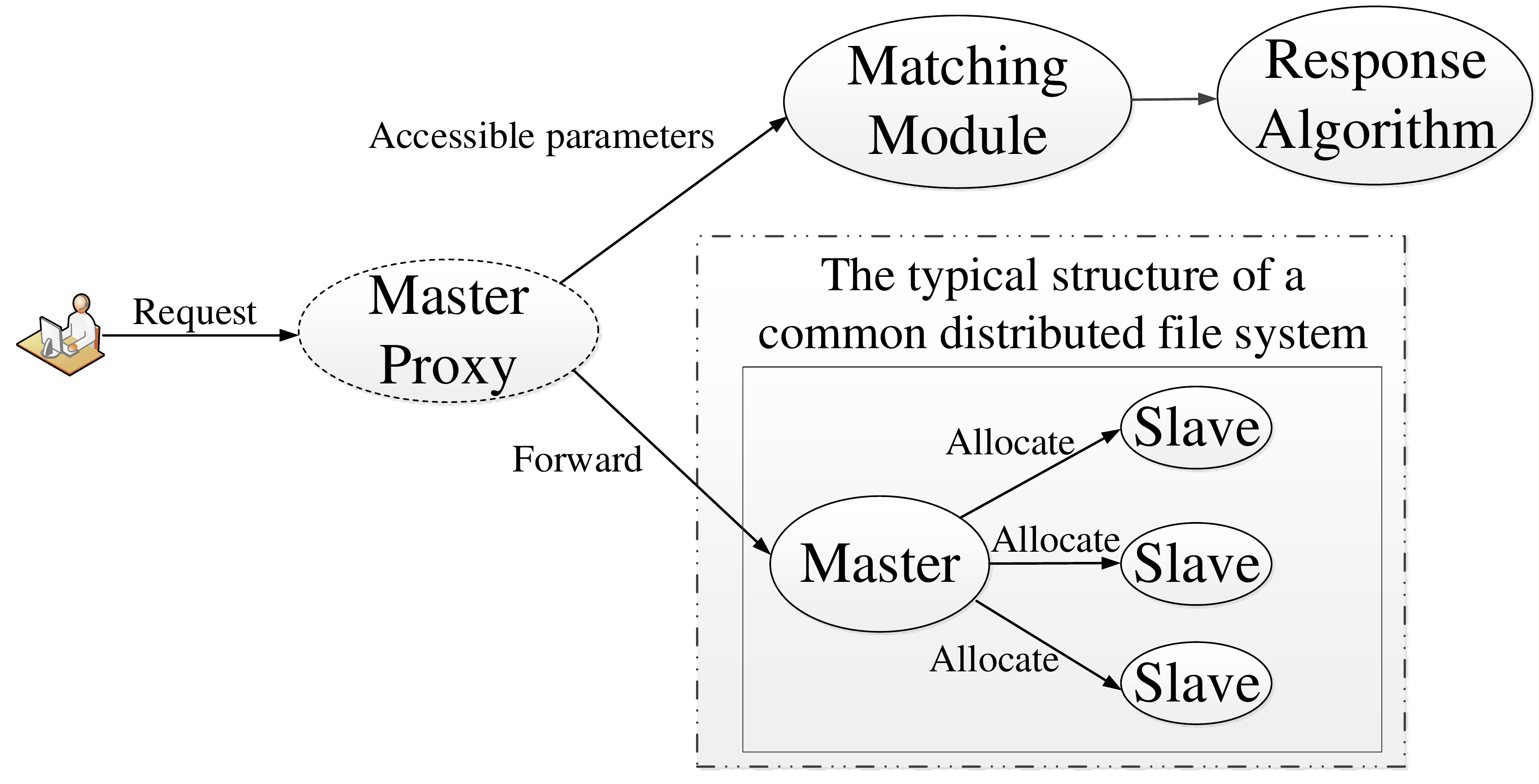}}
   %%The overview of our large-scale image retrieval system.
\end{center}
   \caption{The model of DICP.}
\label{fig:long} \label{fig:onecol}
\end{figure}

The typical structure of a common distributed system provides no specific image processing operation, let alone an effective mechanism. As depicted in the bottom part of Fig. 7, one Master and numerous Slaves play the main roles in the traditional distributed file system where Master accepts the requests from the client and then allocates these requests to those Slaves. Owing to this working mechanism, the new-coming requests that demand urgent processing can not get immediate response. To this end, we design an effective distributed processing mechanism which is implemented by working with the inherent working procedure of a common distributed file system in parallel. The core design of DICP is depicted in the top part of Fig.7, emphasized in Fig. 8. In our DICP mechanism, we design a Master Proxy and a Matching Module on the basis of traditional distributed file system, which cooperate to provide high stability and pressure resistance. Algorithm 2 presents the main implementation procedure of DICP.
\begin{figure}[h]
\begin{center}
%\fbox{
\scalebox{1.3}[1.3]{\includegraphics[width=0.8\linewidth]{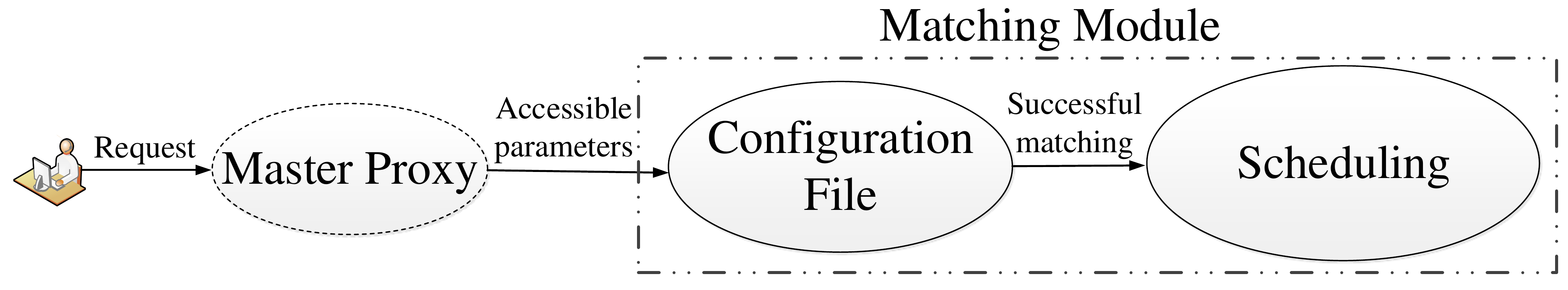}}
   %%The overview of our large-scale image retrieval system.
\end{center}
   \caption{The core design of DICP.} \label{fig:long}
\label{fig:onecol}
\end{figure}

\begin{algorithm}
\caption{DICP Mechanism}
\label{alg1}

\begin{center}\textbf{~~/* \emph{Step 1: Master Proxy} */}\end{center}
\begin{algorithmic}[1]
%\STATE P-Image: an image representation containing filename, width-height, and pixel values
%\STATE Input: Image[1\dots N]
\FOR{\textbf{each} request from the client}
  \STATE  The main thread transmits the request to the Master;
  \STATE  Start a vice thread to extract the parameters from the request and  transfer the parameters to the Matching Module;
  \ENDFOR  \vspace{2ex}

  \begin{center}\textbf{/* \emph{Step 2: Matching Module} */}\end{center}

FN: the pre-set image filename in Matching Module;

FE: the pre-set image extension in Matching Module;
%\algstore{bkbreak}
\FOR{\textbf{each} parameter from Master Proxy}
  \STATE  Extract image filename and extension from the parameters;
  \STATE  Gain FN and FE from the configuration file;
  \IF{parameter.filename$=$FN \textbf{and} parameter.extension$=$

  FE}
  \STATE Start a new thread and call the corresponding algorithm;	
  \ENDIF
  \ENDFOR
\end{algorithmic}
\end{algorithm}

\textbf{Step 1: Master Proxy} (Line 1 to 4). The Master Proxy accepts the requests from the client, and then starts the main thread to transmit these requests to Master to obtain other necessary processing information that the traditional distributed system can provide. In parallel, a vice thread will be started to extract the parameters (the image filename and extension) from the request and transfer the parameters to the Matching Module.

\textbf{Step 2: Matching Module} (Line 5 to 11). When the parameters have been transmitted into the Matching Module, multiple threads will be started to extract the image filename and extension from these parameters to be matched with the configuration information that have been pre-set in the configuration file (Line 5 to 7). If the accessible parameters successfully match with those set beforehand according to real applications, then, the scheduling mechanism would call the related response algorithms in computer vision area, such as image detection, image retrieval, image classification, \begin{slshape}etc..\end{slshape} Then, the processing information provided by the Master-Slave mechanism will be used to collaborate with the  response algorithms to accomplish corresponding requests. For instance, when the client of the mobile terminal asks for image retrieval, Master Proxy will transmits the requests to the Master-Slave and our designed Matching Module. In Matching Module, the accessible parameters refined from the requests will be matched with that set in the configuration file beforehand, and if successfully matched, the scheduling mechanism will call the image processing algorithms (\emph{e.g.} [13][31]) that have been defined in the configuration file to accomplish image retrieval with the processing information provided by Master-Slave. Otherwise, if the parameters fail to be matched, no response will be returned to the client. From this perspective, DICP owes its immediate response to the particular design of the parallel processing modules: one is used to gain the processing information, while the other determines what operation should be implemented.

Note that it is exactly the multithreading design of DICP that greatly enhances the parallelism and hence successfully implements the robust pressure resistance rather than an overload of the cluster on the dynamic input even with a heavy traffic. The experimental results presented in Section 6.5 would prove the stability and pressure resistance of DICP.

\section{Experiments}

This section provides comprehensive experimental evidence for the performance of our proposed ICP: 1) to validate the efficiency of Big-Image over the traditional small image files when acting as input; 2) to verify the time efficiency of SICP when processing large-scale static image data; 3) to prove the stability and pressure resistance of DICP when processing the dynamic input.

\subsection{Experimental Environment}

We provide representative results achieved on the challenging ImageNet [30] dataset running on Hadoop-1.0.3 cluster of two IBM minicomputers, each of which equips with 16-core 2.2GHz IBM CPU and 30GB memory space. Since the hardware architecture of IBM minicomputer is ppc64 bits, we used Java6 SDK also released by IBM to perform better compatibility. The operating system of the two minicomputers is SuseLinuxEnterprise11. In order to bring the Hadoop cluster into full play, we set mapre.map.tasks as 24 and mapred.reduce.tasks as 8, both of which are key setup parameters of Hadoop. In our cluster, the number of Map Node is 8. 

%\begin{figure}[h]
%\begin{center}
%\fbox{
%\scalebox{0.8}[0.8]{\includegraphics[width=0.8\linewidth]{fig88.pdf}}
   %%The overview of our large-scale image retrieval system.
%\end{center}
  % \caption{Experimental environment (left: the front view of two minicomputers;
   %right: the back view of two minicomputers in the machine cabinet )} \label{fig:long}
%\label{fig:onecol}
%\end{figure}

\subsection{ImageNet Dataset}

ImageNet [30] is a large-scale image dataset aiming to provide researchers an easily accessible image database and it is organized according to the WordNet hierarchy. Each meaningful concept in WordNet, possibly described by multiple words or word phrases, is called a “synonym set” or “synset”. There are more than 100,000 synsets in WordNet, majority of which are nouns (80,000+). In ImageNet, approximately 1,000 images are provided to illustrate each synset, and images of each synset are quality-controlled and human-annotated. In its completion, ImageNet will offer tens of millions of cleanly sorted images for most of the synsets in the WordNet hierarchy.

In our aforementioned three experiments, we chose 10,000 images of the same resolution $640 \times 480$ as the dataset denoted by ImageNet-B to validate the efficiency of Big-Image as an input file; 1,000,000 images of different categories as a dataset represented by ImageNet-S to verify the processing efficiency of SICP; and 200 images resized to no larger than $800 \times 800$ indicated by ImageNet-D to test the stability and pressure resistance of DICP. Fig.9 shows the sample images from the ImageNet dataset. Here, ImageNet-B, ImageNet-S and ImageNet-D are independent.

\begin{figure}[h]
\begin{center}
%\fbox{
\scalebox{1.0}[1.0]{\includegraphics[width=0.8\linewidth]{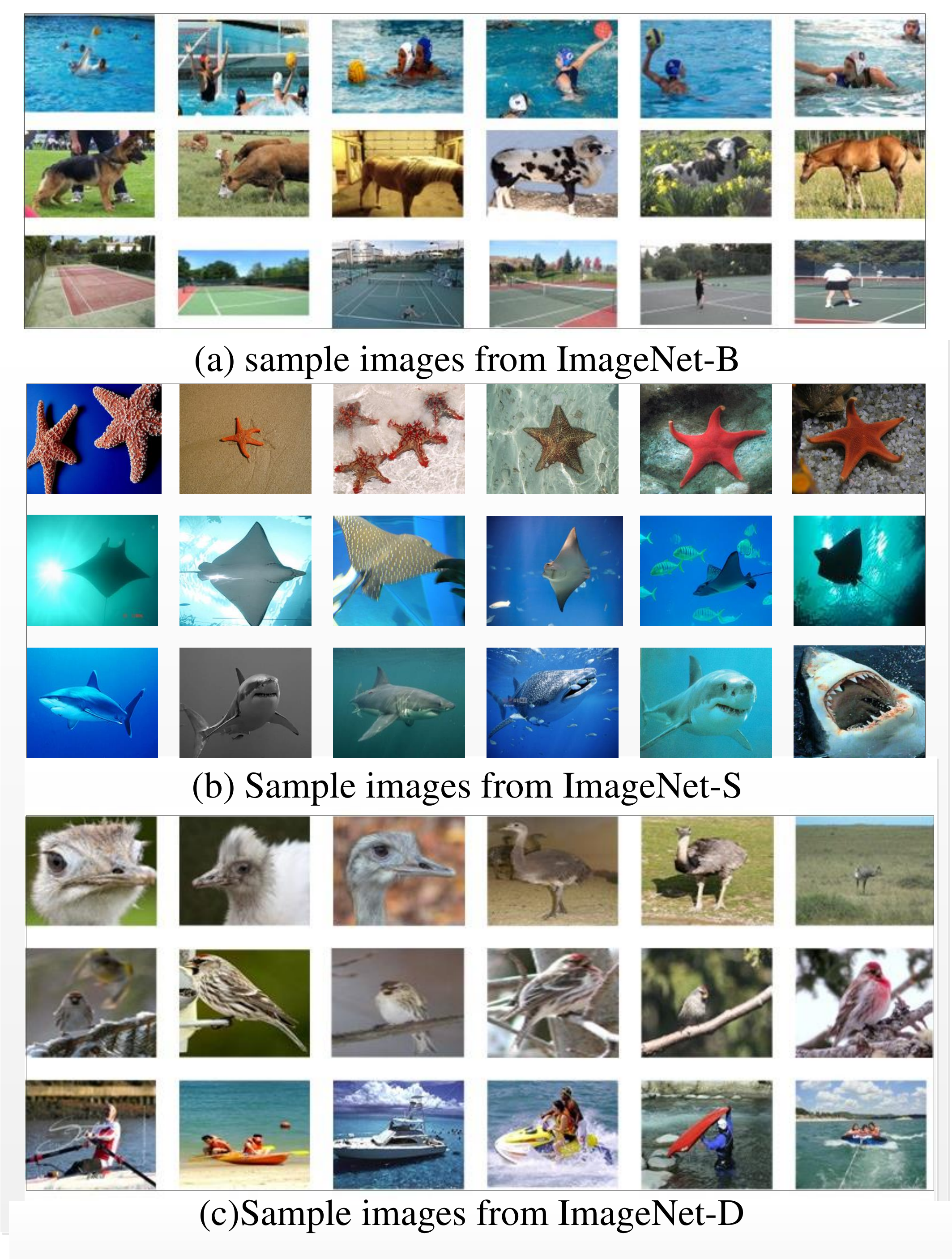}}
   %%The overview of our large-scale image retrieval system.
\end{center}
   \caption{Sample images from ImageNet.} \label{fig:long}
\label{fig:onecol}
\end{figure}

Note that the images from ImageNet-B must be kept the same resolution so as to guarantee a fair condition when comparing the input efficiency between Big-Image and traditional small image files. Meanwhile, in order to exactly infer the disparity in efficiency when separately regarding Big-Image and traditional small image files as input, sufficient images need to be contained and here, experimental setting is 10,000. Similarly, since that SICP is aimed at boosting the time efficiency in processing big image data, the image scale of ImageNet-S should reach a relatively high level so as to persuasively validate the superiority of SICP over traditional methods based on a single node. Here, we choose 1,000,000 images to provide convincing results. As for ImageNet-D, we initially planed to contain 1,000 images, yet seriously restricted by the number of our experimental devices for the moment. Hence, we finally chose 200 images as ImageNet-D and each image in ImageNet-D was resized to no larger than $800 \times 800$ to serve as the dynamic input processed by DICP mechanism. It deserves to be mentioned that the scale of ImageNet-D is plausible to reflect the DICP mechanism to some extent, and the scale is expected to be increased as the experimental environment gets improved, which will be convinced in our future work.

\subsection{Efficiency of Big-Image as An Input File}

In this part, we utilize ImageNet-B to compare the input efficiency between Big-Image and traditional small image files. Each time, the number of input images without any preprocessing is equal to that contained in a single Big-Image. As depicted in Fig. 10, the line marked with squares stands for the growth tendency of traditional small image files' input time, while the one marked with diamonds is for Big-Image.
 Here, we only count the input time (input Big-Image into the devices), not including the consumption of producing P-Image and Big-Image, 
 the reason for which has been discussed at the end part of Section 4.1. The input cost of different image 
 amount is labeled next to the corresponding coordinate point. For instance, (4,935) represents the time cost 
 of inputting 10,000 traditional small image files. From Fig. 10, it is hard to tell any difference from the 
 two smooth linearity curves when the input number is limited to less than 500. However, once the input amount 
 is larger than 500, subtle difference starts to appear especially when the image scale exceeds 1,000. When inputting 1,000 images, 
 the input consumption of small image files is 562 seconds, 1.6 times of that of Big-Image which only costs 352 seconds. In Fig. 10, 1,000 images can be regarded as a turning point, after which the curve representing traditional small images shows an exponential increase while the curve representing single Big-Image remains a smooth and gently linear growth. When inputting 2,000 images, the input time of small image files is 1,093 seconds, nearly 1.83 times of that of Big-Image which only costs 598 seconds. When increasing the image amount to 4,000, the input consumption of small image files is nearly 3.18 times of that of Big-image. Once the image scale reaches to 10,000, the input time of small image files soars to 4,935 seconds while Big-Image only costs 780 seconds, the comparison is about 6.33 times. According to the growth tendency of the two curves, we can infer that when the scale of input images soars to a certain level, the superiority of our Big-Image will be revealed more obviously. The conspicuous contrast of efficiency between the two forms of input owns to the well-designed processing mechanism of Big-Image. Considering the input of traditional small image files, once the input images reach such an extensive scale that the cluster just cannot process them timely, then the later input must be delayed, which leads the input time to gain a sharp boost just as presented in Fig. 10. By contrast, our well-designed Big-Image successfully releases this constraint by merging the small image files into a large one and then partitioning the large file into several cells to be processed in parallel. Thus, it equals to process several files no matter how many images are input. From this perspective, the increase of image scale has little influence on the efficiency of Big-Image when used as input file.

\begin{figure}[h]
\begin{center}
%\fbox{
\scalebox{0.9}[0.9]{\includegraphics[width=0.8\linewidth]{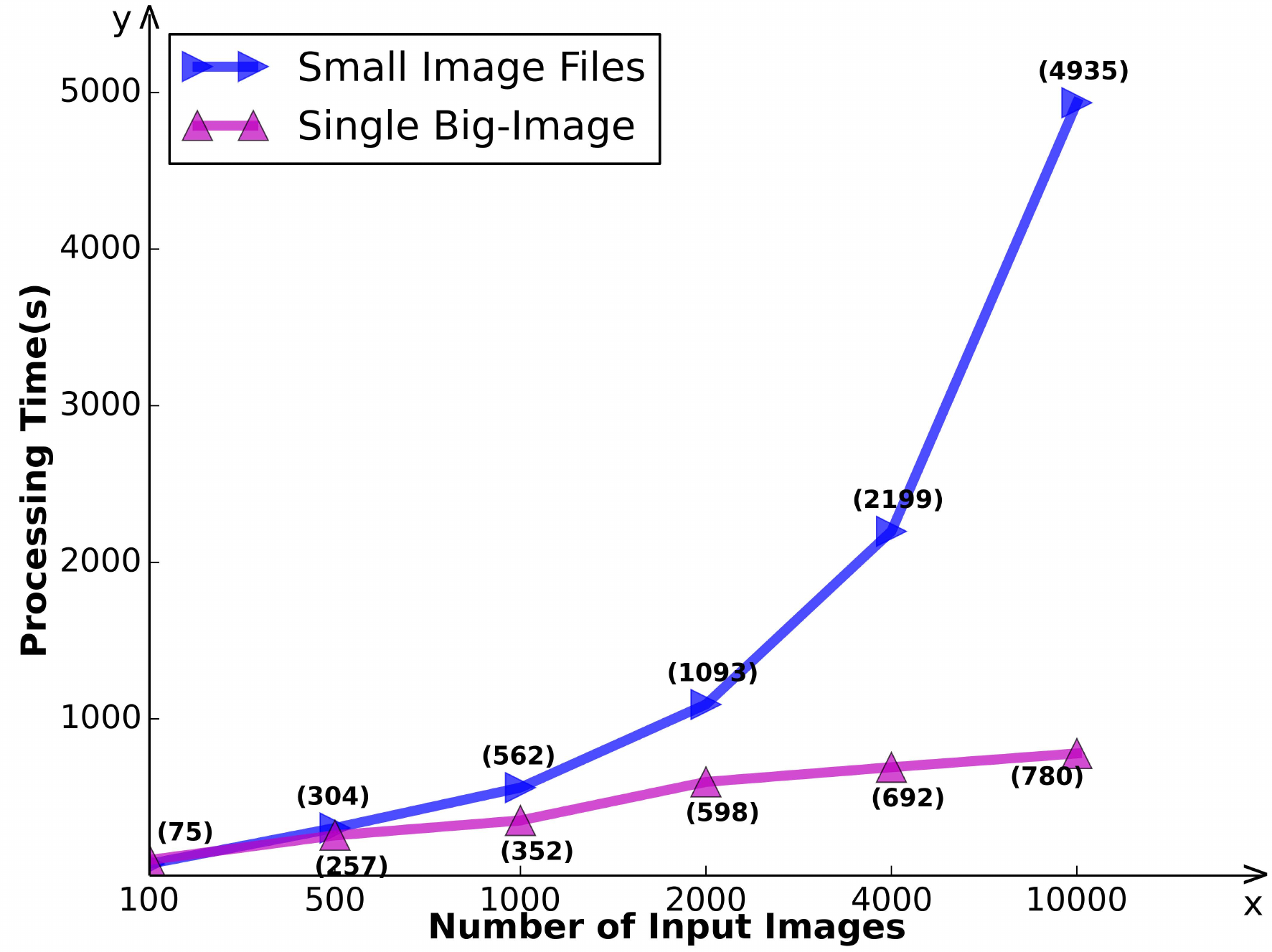}}
   %%The overview of our large-scale image retrieval system.
\end{center}
   \caption{The comparison of efficiency between Big-Image and traditional
   small image files as input.
} \label{fig:long} \label{fig:onecol}
\end{figure}

\subsection{Time Efficiency of SICP}

In this section, we aim to validate the time efficiency of SICP by using ImageNet-S to implement two classical 
image processing algorithms, \emph{i.e.} Harris [31] and SIFT [13], both on ICP and OpenCV. OpenCV is a prevalent 
framework also designed for improving computational efficiency and with a strong focus on real applications. 
Therefore, we employ OpenCV to represent the traditional image processing method based on a single node. 
To guarantee a fair comparing condition, we use two machines to implement OpenCV, which cooperate to process 
the same image data as SICP framework. Here, considering the performance of the cluster, we set the NumMapTask as 32, and the BLOCKSIZE is accordingly 3200MB. From the results obtained from ICP and OpenCV, time efficiency of SICP 
would gain creditable verification.

\subsubsection{Harris}

The Harris [31] corner detector is a popular interest point detector due to its strong invariance to rotation, scale, illumination variation and image noise. Based on the local auto-correlation function, Harris is utilized to cater for image regions containing texture and isolated features. Since the algorithmic complexity of Harris is at a relatively low level, it is efficient enough to process small-scale images on a single machine. However, despite the simpleness of image algorithms such as Harris, the requirement for time efficiency based on a single node is hard to meet along with the increasing scale of big image data. Our SICP is proposed to release this restriction, which successfully achieves high efficiency when processing the big image data stored in the distributed system. Table 1 records the experimental results of SICP and OpenCV when processing different scales of images.

\begin{table}[h]
\centering \caption{The comparison between SICP and OpenCV on
Harris.}
\begin{tabular}{|c|c|c|c|c|c|}
\hline \#Image&10,000&50,000&100,000&500,000&1,000,000\\
\hline OpenCV(min)&7.91&37.62&83.69&419.72&866.79\\
\hline SICP(min)&12.81&27.47&49.02&79.62&148.28\\ \hline
\end{tabular}
\end{table}

From the experimental results recorded in Table 1, when the number of images is only 10,000, the processing time of SICP is 12.9 minutes, 4.89 minutes longer than that of OpenCV. The rationale for this unexpected result owns to the low complexity of Harris which can be implemented efficiently even on a single node. Recall what has been elaborated in Section 4, the processing mechanism of SICP requires some preprocessing such as decoding the images to get P-Image, the mergence and segmentation of Big-Image, \begin{slshape}etc..\end{slshape} Time cost of these necessary operations accounts for a large proportion in the total cost of small-scale images processing. Therefore, the superiority of SICP over traditional methods based on a single node is hard to demonstrate when the image amount is limited to a low level, which, however, can be desirably solved by employing our DICP mechanism. Table 1 gives a desirable comparison when the number of images reaches 50,000, where the processing time of SICP is 10.15 minutes less than that of OpenCV. Along with the increasing number of images, we can validate the efficiency of SICP through a simple calculation. For example, time cost of OpenCV is 1.7 times of SICP's when processing 100,000 images, and this comparison reaches 5.3 times when processing 500,000 images. Furthermore, when the scale of images boosts to 1,000,000, we can see a comparison of 5.8 times. Based on this growth tendency, we can expect that our ICP framework would undoubtedly outperform OpenCV, whatever the complexity of algorithms, provided that the number of images is large enough.

\subsubsection{SIFT}

SIFT[13] (Scale-invariant feature transform) is a popular algorithm for extracting distinctive invariant features from images that can be used to perform reliable matching between different views of an object or scene. The features are invariant to image scale and rotation and are shown to provide robust matching across a substantial range of affine distortion, change in 3D viewpoint, addition of noise, and change in illumination. As one of the most classical local feature algorithms in recent 10 years, SIFT not only contributes a lot to the academic community but also gains the acceptance from the industries in computer vision, machine learning, graphics, and other related areas. Compared with Harris, the algorithmic complexity of SIFT is much higher, which results in a larger time consumption even when processing small-scale images on a single node. Owing to this, the aforementioned time cost of the necessary preprocessing of SICP only accounts for a little proportion in the whole processing time. Hence, SICP outperforms OpenCV by a large margin even when processing small-scale images. Table 2 gives the experimental results of SICP and OpenCV when separately implementing SIFT.

\begin{table}[h]
\centering \caption{The comparison between SICP and OpenCV on SIFT.}
\begin{tabular}{|c|c|c|c|c|c|}
\hline \#Image&10,000&50,000&100,000&500,000&1,000,000\\
\hline OpenCV(min)&181.87&1012.80&2065.69&10439.22&20962.52 \\
\hline SICP(min)&20.11&102.52&203.63&913.05&1281.80\\ \hline
\end{tabular}
\end{table}

Just as recorded in Table 2, time cost of OpenCV when processing 10,000 images is 9.09 times of that when using SICP, and this comparison reaches more than 9.879 times when the number of images is 50,000. When the image scale boosts to 100,000 and 500,000, time consumption of OpenCV is more than 10 and 11.4 times of SICP, respectively. Once the number of images soars to 1,000,000, we can observe 16.35 times in processing cost that OpenCV exceeds SICP. The analyzed experimental results have powerfully validated the increasing superiority of SICP over OpenCV along with the growing image scale.

According to the analysis of Table 1 and Table 2, we can draw the conclusion that both of the algorithmic complexity and the image scale can influence the performance of SICP while the image scale plays the dominate role. Despite the low algorithmic complexity, high efficiency of SICP over the conventional methods based on a single node can be apparently revealed as long as the number of images is large enough. Considering the practical applications in image processing field, the complexity of image processing algorithms is usually higher than Harris and SIFT, and besides, the image scale is much larger than that contained in our ImageNet-S. Hence, we can obviously obtain high efficiency when employing SICP to implement image processing algorithms that involve big image data stored in the distributed system. The real applications have not been evaluated due to the experimental environment and data amount.  However, Google \begin{slshape}etc.\end{slshape} industry giants have been successfully processing an increasing number of big data on MapReduce, which largely benefits from the inherent scalability that MapReduce possesses. In this sense, it can be scalable to accomplish those demanding tasks in real applications with our framework.

\subsubsection{Comparison of Visual Results}

We have verified the time efficiency of SICP in Section 6.4.1 and 6.4.2, yet another factor that also counts a lot in the performance evaluation of ICP framework draws our attention, \emph{i.e.}, the visual results of local features. Fig. 11 and Fig. 12 presents some example images of our experimental results on Harris and SIFT, respectively. Each couple of identical images separately processed by ICP and OpenCV are regarded as one group.

\begin{figure}[h]
\begin{center}
%\fbox{
\scalebox{1.2}[1.2]{\includegraphics[width=0.8\linewidth]{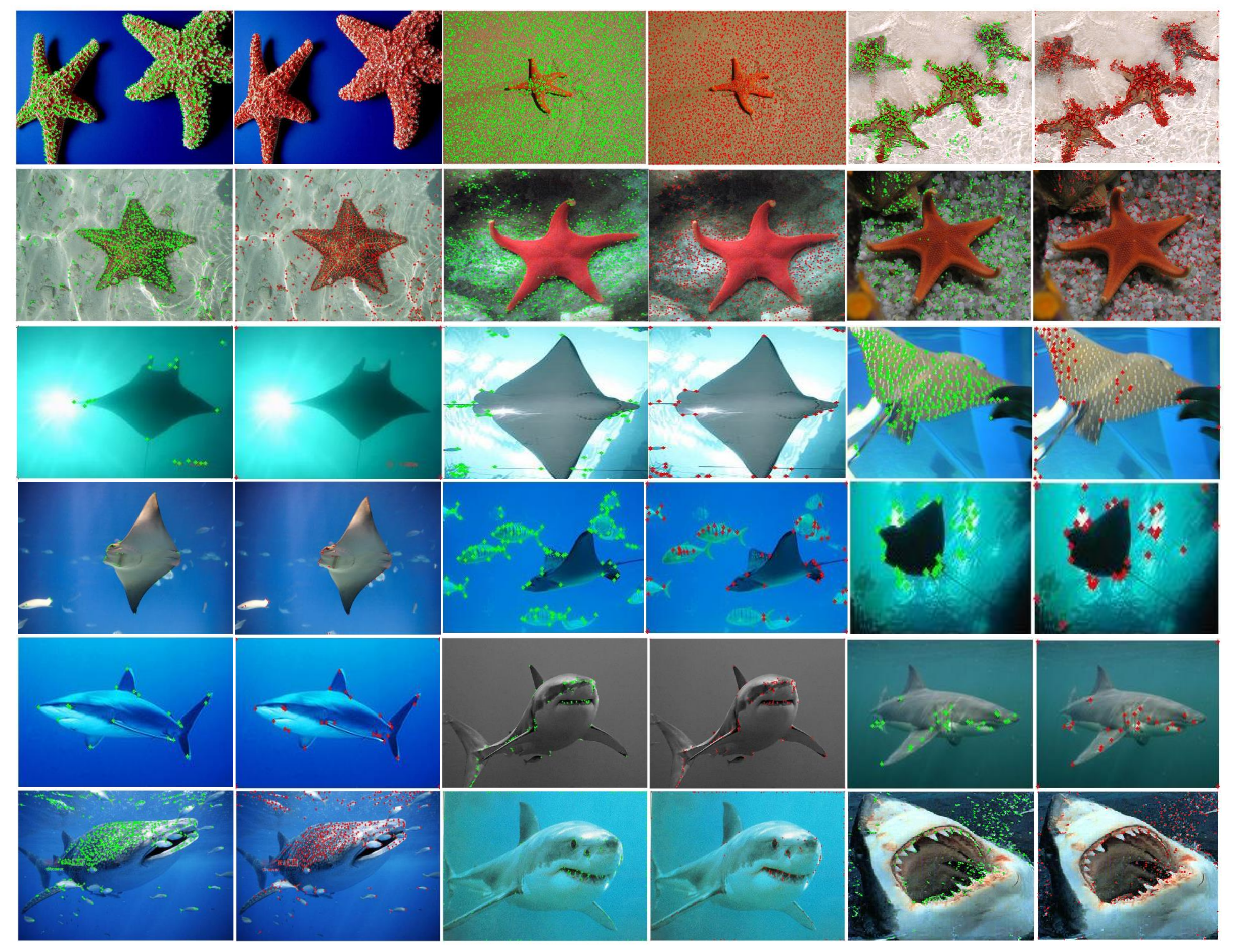}}
   %%The overview of our large-scale image retrieval system.
\end{center}
   \caption{Examples of visual experimental results on Harris. The
odd and even number of columns present the images processed on ICP
and OpenCV, respectively.} \label{fig:long} \label{fig:onecol}
\end{figure}

\begin{figure}[h]
\begin{center}
%\fbox{
\scalebox{1.2}[1.2]{\includegraphics[width=0.8\linewidth]{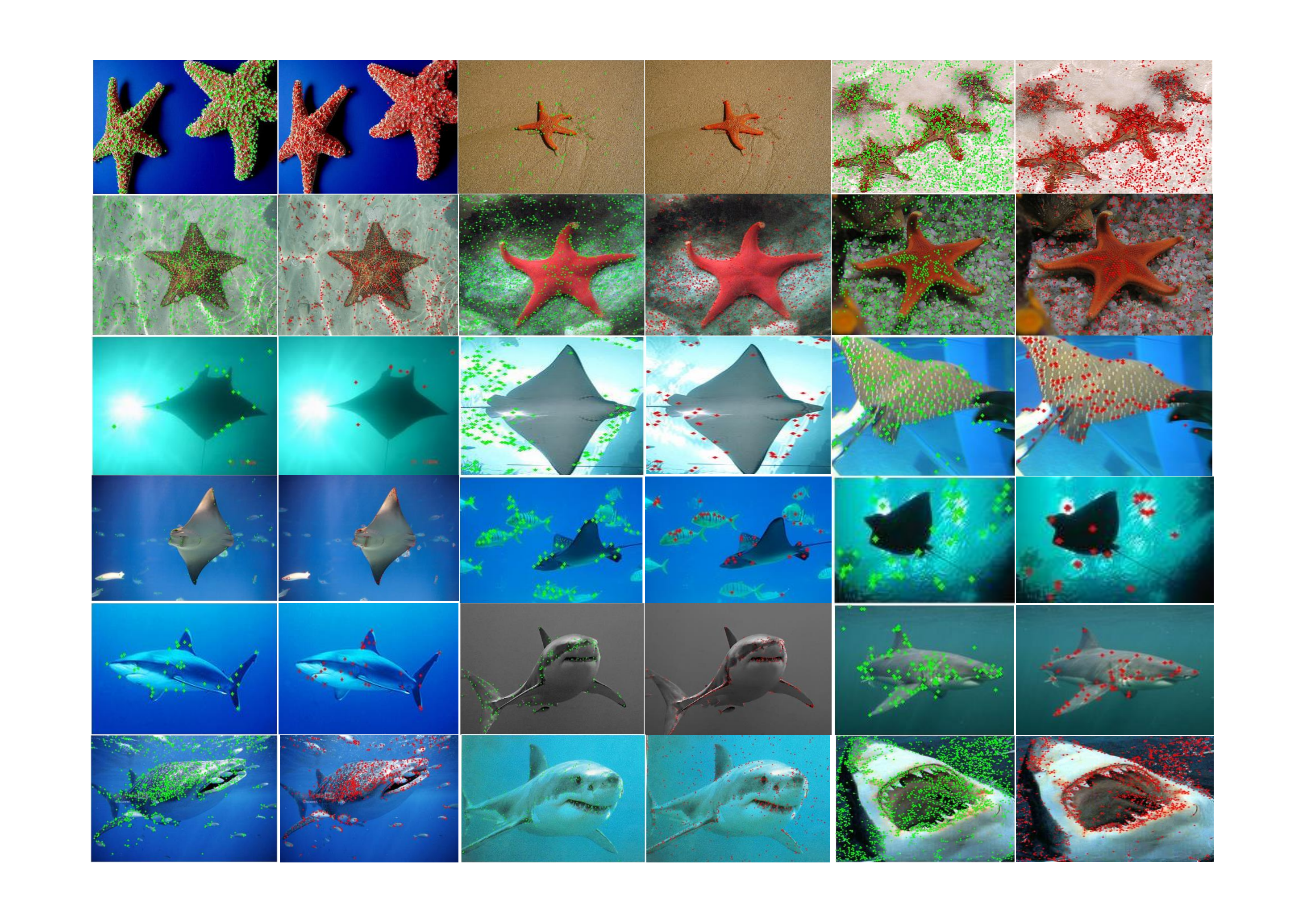}}
   %%The overview of our large-scale image retrieval system.
\end{center}
   \caption{Examples of visual experimental results on SIFT. The
odd and even number of columns present the images processed on ICP
and OpenCV, respectively.} \label{fig:long} \label{fig:onecol}
\end{figure}

From Fig. 11, we can clearly observe that both ICP and OpenCV have successfully detected the interest points by implementing Harris, while ICP obviously performs more plausibly from the perspective of subjective assessment. For instance, the left twin images in the \begin{math}1^{st}\end{math} row shows the similar number of interest points while the left group listed in the \begin{math}3^{rd}\end{math} row presents different results where ICP has apparently detected more interest points than OpenCV. Other groups of twin images in Fig. 11 can also prove the superiority of our ICP over OpenCV when considering both time efficiency and local visual results. Fig. 12 provides another convincing instance where the left twin images presented in the \begin{math}1^{st}\end{math} row show similar results while the middle group in the bottom row presents much difference in which ICP extracts feature points further more accurately than OpenCV via subjective evaluation. Similarly, other groups listed in Fig. 12 demonstrate the better performance of ICP over OpenCV in different degrees when implementing SIFT. According to the aforementioned analysis, we can conclude that though our SICP is aimed at boosting the time efficiency of processing big image data, it does not compromise the quality of the visual processing results and the performance of procedure in between. From this perspective, our ICP framework outperforms traditional processing methods based on a single node such as OpenCV in significant measure.

\subsection{Performance of DICP}

Just as mentioned in Section 5 , the key success of DICP mainly depends on its stability and pressure resistance. Hence, in this part, we would like to present two types of experiments using our ImageNet-D to separately evaluate the stability and the pressure resistance of DICP. Here, the performance of DICP is also validated via Harris and SIFT algorithms.

\subsubsection{Stability of DICP}
Here, we define the stability of our DICP mechanism as follows: Each time, no matter how many small-scale and new-coming images need to be processed, the average processing time of each image is always in a little range. To test the stability of DICP, we uploaded a small number of images in series. Specifically, we uploaded several images each time, and this operation would be repeated for 10 times with little time slot. Here, we uploaded the images for two turns. For the first turn, we regularly uploaded 10 images each time, and for the second, the number of uploaded images each time was random but always no more than 10.

\begin{figure}[h]
\begin{center}
%\fbox{
\scalebox{1.25}[1.25]{\includegraphics[width=0.8\linewidth]{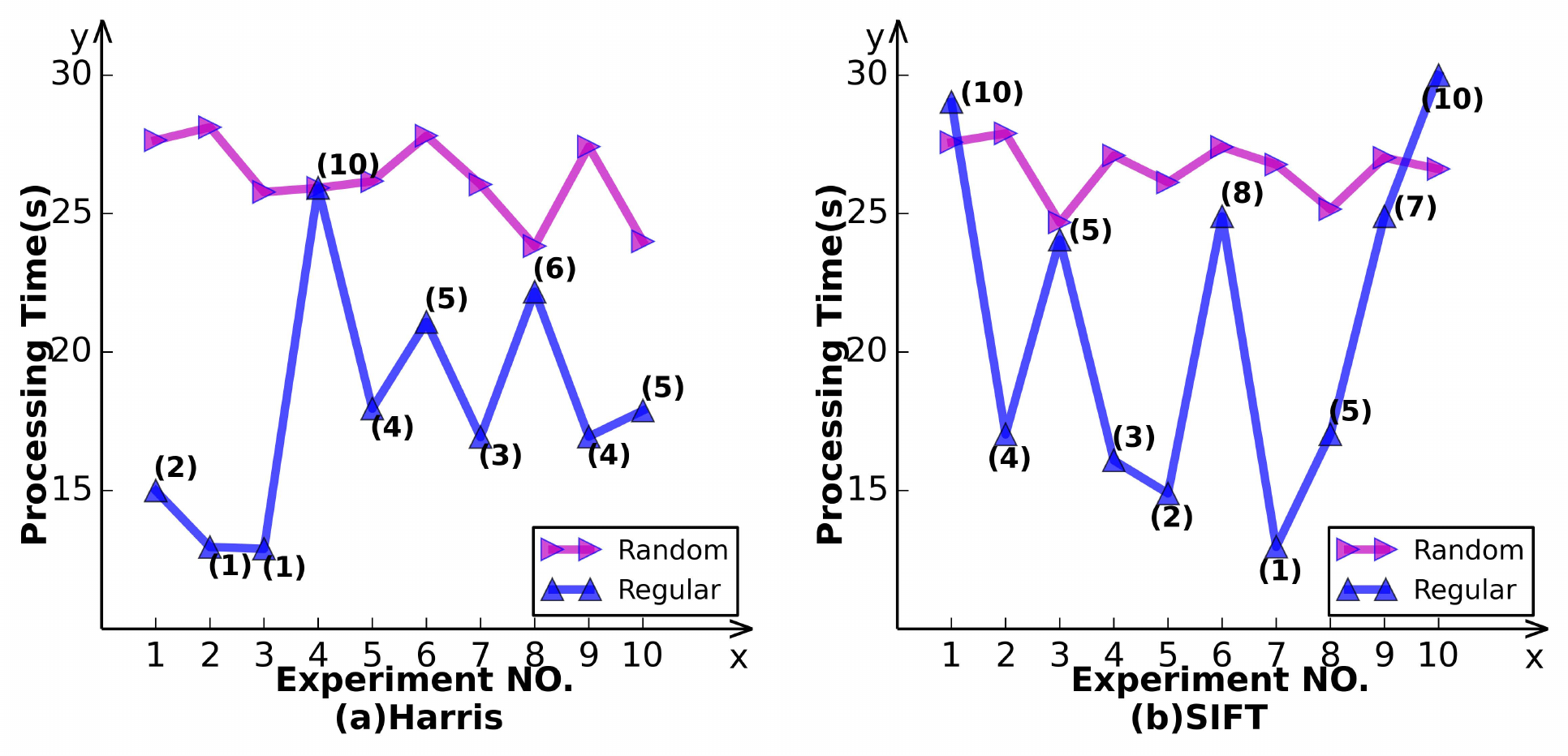}}
   %%The overview of our large-scale image retrieval system.
\end{center}
   \caption{Stability test on DICP with different algorithms.}
\label{fig:long} \label{fig:onecol}
\end{figure}

Fig. 13 illustrates the experimental results of Harris and SIFT on DICP; In Fig. 13, the curve marked with diamonds represents the processing time of regular 10 images; the curve marked with squares stands for the processing time of a random number of images, and the value in the brackets labeled next to each square is the corresponding image amount. Observing from (a) and (b) in Fig. 13, when the number of images is fixed to 10, the processing cost of each time always differs less than 5 seconds, which illustrates that our DICP mechanism can work steadily when the image amount is fixed. Taking a closer look at the results of the random input, we can validate the stability of our DICP after a simple calculation. In (a), we can calculate the average processing time (measured in seconds) of each case: 7.5, 8, 8, 2.7, 4, 4.4, 5.3, 3.8, 4.3, 3.6. The dissimilarity between the highest and lowest time is only 5.3 seconds. Another persuasive argument draws our attention to (b). After a simple calculation, we get the average processing time (measured in seconds) of each case: 2.9, 4.1, 4.8, 5.3, 7.5, 3.125, 7.5, 3.3, 3.6, 3. The disparity between the highest and lowest time is only 4.6 seconds. By analyzing the results gained from the regular and random input, we have successfully validated the stability of our DICP mechanism.

\subsubsection{Pressure Resistance of DICP}
Pressure resistance represents the load capacity of our ICP framework. To prove the pressure resistance of DICP, we constantly uploaded the 200 images from ImageNet-D to observe if the cluster would fail to timely process these continuous input and result in a breakdown. The correspondence between the processing time and the completion rate is clearly depicted in Fig. 14. As is known to all, the procedures of image input and image processing are parallel, while the time cost of input is much less than that of processing. Due to this, once the input images fail to be processed in time, the whole framework would possibly gain an undesirable breakdown. In our experiment, we continually input the images from ImageNet-D to be processed, which have already overstepped the load limit of the cluster, yet no breakdown came along. Instead, just as depicted in Fig. 14, the processing time and the completion rate show a linear relationship, and it is exactly the linear increase that powerfully validates the excellent pressure resistance of our DICP due to the good ability of its internal scheduling described in Section 5. According to Fig. 14, we can infer that if we can successfully upload 1,000 images without the limitation of experimental devices, we can get the similar linear increase just like that depicted in Fig. 14, whereas the total processing time would undoubtedly enhance. From this perspective, as long as the cluster is large enough to support the large number of input images, our DICP mechanism can still show its robust pressure resistance. While the superiority of DICP is hard to reveal once the image scale is too large due to the limited devices, our SICP mechanism is ready to show its good performance.

\begin{figure}[h]
\begin{center}
%%\fbox{
\scalebox{0.9}[0.9]{\includegraphics[width=0.8\linewidth]{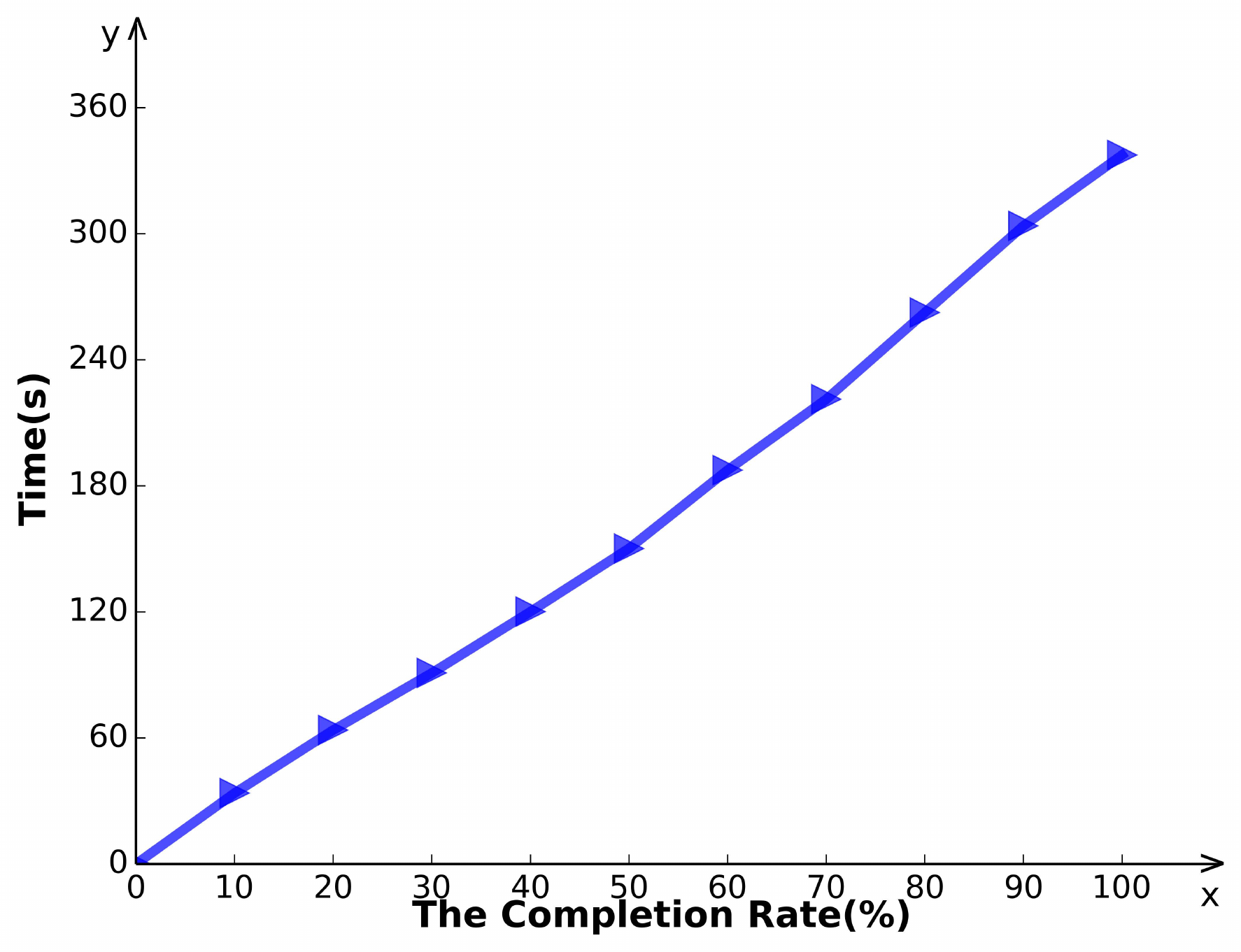}}
   %%The overview of our large-scale image retrieval system.
\end{center}
   \caption{Pressure resistance test on DICP.} \label{fig:long} \label{fig:onecol}
\end{figure}

\subsection{Discussion}

In aforementioned experiments, performance of our proposed ICP framework 
has been powerfully validated on the Hadoop cluster. 
It deserves to be mentioned that our ICP framework is mainly aimed at improving processing efficiency, 
so we choose to compare with the traditional image processing framework based on a single node, \emph{e.g.} OpenCV, 
a prevalent framework also designed for improving computational efficiency and with a strong focus on real 
applications. As two novel data representations, P-Image and Big-Image show their superiority in minimizing 
the input consumption just as depicted in Fig.10, which contributes a lot to reduce the whole image processing 
period when compared with traditional image processing approaches. Since that the procedure of producing P-Images 
and Big-Image needs certain slot cost, the image scale must be large enough to show the advancement of SICP on the 
I/O efficiency over traditional small image files, which, however, needs no doubt because our SICP is actually aimed 
at processing big image data. And on the other hand, just as discussed in the end part of Section 4.1, the image processing,
 P-Image producing, and Big-Image producing can be implemented in parallel, which demonstrates that producing P-Image and Big-Image 
 will not drag down the whole efficiency. The experimental results presented in Table 1 and Table 2 provide representative instances
 to verify the time efficiency of SICP in processing images with different algorithms. Low-complexity algorithms and small-scale images 
 would make serious constraints on the performance of SICP, which might be even inferior to traditional methods based on a single node.
 Again, since our SICP is employed to process large-scale images in the cloud computing manner, no matter the complexity of image 
 processing algorithms or the image amount will be far higher than that in our experiment. The boost of time efficiency does not 
 mean that we ignore the quality of the final processing results, which can be certified through the example results presented in
 Fig.11 and Fig.12. Just as Fig. 13 and Fig. 14 have validated, DICP owns high stability and pressure resistance due to its
 effective internal scheduling mechanism, whereas it is probably the scheduling mechanism that might affect the performance 
 of DICP, say, waiting for the query usually costs most of the time. We have been devoted to proposing a comprehensive scheduling
 mechanism that can achieve both robust pressure resistance and low time cost. According to the experimental evaluation, 
 our ICP framework has gained significant advancements in boosting the time efficiency of image processing,
 and our efforts may perfect this target in the near future.

\section{Conclusion}
This paper elaborates an effective distributed processing framework named ICP aiming to efficiently process the large-scale image data without compromising the results quality. ICP contains two types of processing mechanism, \emph{i.e.} SICP and DICP, to achieve effective processing on the static big image data and the dynamic input, respectively. Collaborating with MapReduce, P-Image and Big-Image play the key roles of SICP to boost the time efficiency. Relying on the two newly proposed structures, time efficiency would be greatly improved by utilizing SICP to process large-scale images stored in the distributed system when compared with traditional methods based on a single node. If the new-coming image files need to be processed urgently, DICP allows for immediate response
without any delay to avoid undermined problems. Extensive experiments have been conducted on ImageNet dataset to validate the efficiency of ICP. From the desirable results, we believe that big image data processing is a promising direction, which calls for endeavor in infrastructure, computing framework, modeling, learning algorithm, applications, and all walks of life.

\appendices

% you can choose not to have a title for an appendix
% if you want by leaving the argument blank

% use section* for acknowledgment
\ifCLASSOPTIONcompsoc
  % The Computer Society usually uses the plural form
  \section*{Acknowledgments}
\else
  % regular IEEE prefers the singular form
  \section*{Acknowledgment}
\fi

This work was supported in part by the National Natural Science Foundation of China under Grant 61370149, in part by the Fundamental Research Funds for the Central Universities(ZYGX2013J083), and in part by the Scientific Research Foundation for the Returned Overseas Chinese Scholars, State Education Ministry.

% Can use something like this to put references on a page
% by themselves when using endfloat and the captionsoff option.
\ifCLASSOPTIONcaptionsoff
  \newpage
\fi

% trigger a \newpage just before the given reference
% number - used to balance the columns on the last page
% adjust value as needed - may need to be readjusted if
% the document is modified later
%\IEEEtriggeratref{8}
% The "triggered" command can be changed if desired:
%\IEEEtriggercmd{\enlargethispage{-5in}}

% references section

% can use a bibliography generated by BibTeX as a .bbl file
% BibTeX documentation can be easily obtained at:
% http://www.ctan.org/tex-archive/biblio/bibtex/contrib/doc/
% The IEEEtran BibTeX style support page is at:
% http://www.michaelshell.org/tex/ieeetran/bibtex/
%\bibliographystyle{IEEEtran}
% argument is your BibTeX string definitions and bibliography database(s)
%\bibliography{IEEEabrv,../bib/paper}

\begin{thebibliography}{50}

\bibitem{IEEEhowto:J. S. Pan}  %1
%This is an example of a book reference
 R. Hong, Y. Yang, M. Wang, X. Hua, ``Learning Visual Semantic Relationships for Efficient Visual Retrieval," \begin{slshape}IEEE Transactions on Big Data,\end{slshape} vol.1, no.4, pp.152-161, 2015.

\bibitem{IEEEhowto:J. M. Guo}    %2
 K. Huang, C. Wang and D, Tao, ``High-Order Topology Modeling of Visual Words for Image Classification," \begin{slshape}IEEE Transactions on Image Processing,\end{slshape} vol.24, no.11, pp.3598-3608, 2015.

\bibitem{IEEEhowto:X. Z. Wen}    %3
 X. Tian, Y. Lu, N. Stender, L. Yang, D. Tao, ``Exploration of Image Search Results Quality Assessment," \begin{slshape}IEEE Transactions on Big Data,\end{slshape} vol.1, no.3, pp.95-108, 2015.

\bibitem{IEEEhowto:X. D. Zhao}    %4
F. Wu, Z. Wang, Z. Zhang, Y. Yang, J. Luo, ``Weakly Semi-Supervised Deep Learning for Multi-Label Image Annotation," \begin{slshape}IEEE Transactions on Big Data,\end{slshape} vol.1, no.3, pp.109-122, 2015.

\bibitem{IEEEhowto:J. Jiang}   %5
Y. Yang, F. Shen, H. T. Shen, H. Li, X. Li, ``Robust Discrete Spectral Hashing for Large-Scale Image Semantic Indexing," \begin{slshape} IEEE Transactions on Big Data,\end{slshape} vol.1, no.4, pp.162-171, 2015.

\bibitem{IEEEhowto:Z. P. Zhang}  %6
Y. C. Wang, C. C. Han, C. T. Hsieh, et al, ``Biased Discriminant Analysis With Feature Line Embedding for Relevance Feedback-Based Image Retrieval," \begin{slshape}IEEE Transactions on Multimedia,\end{slshape} vol.17, no.12, pp.2245-2258, 2015.

\bibitem{IEEEhowto:J. S. Xu}  %7
J. S. Xu, Q. Wu, J. Zhang, F. Shen and Z. M. Tang, ``Boosting Separability in Semisupervised Learning for Object Classification," \begin{slshape}IEEE Trans. Circuits and Systems for Video Technology,\end{slshape} vol.24, no.7, pp.1197 - 1208, 2014.


\bibitem{IEEEhowto:L. Dong}  %dong TIP
L. Dong, J. Su and E. Izquierdo, ``Scene-oriented Hierarchical Classification of Blurry and Noisy Images," \begin{slshape}IEEE Trans. Circuits and Systems for Video Technology,\end{slshape} vol.21, no.5, pp.2534-2545, 2012.

\bibitem{IEEEhowto:L. Dong}  %dong TCSVT
L. Dong and E. Izquierdo, ``A Biologically Inspired System for Classification of Natural Images," \begin{slshape}IEEE Transactions on Image Processing,\end{slshape} vol. 17, no. 5, pp.590-603, 2007.


%10
\bibitem{IEEEhowto:A. Iosup}
A. Iosup, S. Ostermann,  M. N. Yigitbasi,  R. Prodan,  T. Fahringer, D.H.J.Epema, ``Performance analysis of cloud computing services for many-Tasks scientific computing," \begin{slshape}IEEE Trans. Parallel and Distributed Systems,\end{slshape} vol.22, no.6, pp.931-945, 2011.

%11
\bibitem{IEEEhowto:Y. Lin}
Y. Lin, F. Lv, S. H. Zhu, and M. Yang\begin{slshape}et al\end{slshape}, ``Large-scale image classification: fast feature extraction and SVM training," \begin{slshape}CVPR\end{slshape}, pp.1689-1696, 2011.


%14
\bibitem{IEEEhowto:J. Su}
J. Su, L. Dong, P. Ren, and E. R. Hancock, ``Hypergraph matching based on marginalized constrained compatibility," \begin{slshape}ICPR\end{slshape}, pp.2922-2925, 2012.

%15
\bibitem{IEEEhowto:D. G. Lowe}
D. G. Lowe, ``Distinctive image features from scale-invariant keypoints," \begin{slshape} International Journal of Computer Vision\end{slshape}, vol.60, no.2, pp.91-110, 2004.



%17
\bibitem{IEEEhowto: J. Dean}
J. Dean and S. Ghemawat, ``MapReduce: simplified data processing on large clusters," \begin{slshape}Communications of the ACM\end{slshape}, vol.51, no.1, pp.107-113, 2008.

%18
\bibitem{IEEEhowto:X.Y. Zhang}
X. Y. Zhang, L. T. Yang,  C. Liu, J. J. Chen , ``A Scalable Two-Phase Top-Down Specialization Approach for Data Anonymization Using MapReduce on Cloud," \begin{slshape}IEEE Trans. Parallel and Distributed Systems,\end{slshape} vol.25, no.2, pp.363-373, 2014.


%19
\bibitem{IEEEhowto: A. Nandi}
A. Nandi, C. Yu, P. Bohannon, and R. Ramakrishnan, ``Data cube materlization and mining over MapReduce," \begin{slshape}IEEE Transactions on Knowledge and Data Engineering\end{slshape}, vol.24, no.10, pp.1747-1759, 2012.

%20
\bibitem{IEEEhowto: B. Scholkopf}
B. Scholkopf, J. Platt, and T. Hofmann, ``MapReduce for machine learning on multicore," \begin{slshape}NIPS\end{slshape}, pp.281-288, 2007.

%21
\bibitem{IEEEhowto: I. Demir}
I. Demir and A. Sayar, ``Hadoop plugin for distributed and parallel image processing," \begin{slshape}SIU\end{slshape}, pp.1-4, 2012.

%23
\bibitem{IEEEhowto: I. Palit}
I. Palit, and C. K. Reddy, ``Scalable and parallel boosting with MapReduce," \begin{slshape}IEEE Transactions on Knowledge and Data Engineering\end{slshape}, vol.24, no.10, pp.249-255, 2009.


%26
\bibitem{IEEEhowto: IBM}
IBM InfoSphere Streams home page.
http://www-01.ibm.com/software/data/infosphere/streams/

%27
\bibitem{IEEEhowto: Jeff Jonas}
Jeff Jonas, Jonas’s Entity Analytics software home page.
http://www-01.ibm.com/software/data/identity-insight-solutions/

%28
\bibitem{IEEEhowto:F. Chang}
F. Chang, J. Dean, S. Ghemawat, W. C. Hsieh and D. A. Wallach \begin{slshape}et al\end{slshape}, ``Bigtable: a distributed storage system for structured data," \begin{slshape}ACM Transactions on Computer Systems\end{slshape}, vol.26, no.2:4, 2008.

%29
\bibitem{IEEEhowto:S. Ghemawat}
S. Ghemawat, H. Gobioff, and S.-T. Leung, ``The Google file system," \begin{slshape}ACM SIGOPS Operating Systems Review\end{slshape}. vol.37, no.5, pp.29-43, 2003.



%33
\bibitem{IEEEhowto:H. Yu}
H. Y. Yu and D. S. Wang, ``Mass log data processing and mining based on Hadoop and cloud computing," \begin{slshape}ICCSE\end{slshape}, pp.197-202, 2012.



%35
\bibitem{IEEEhowto:J. Zhang}
J. Zhang, G. Q. Wu, X. G. Hu, and X. D. Wu, ``A distributed cache for Hadoop distributed file system in real-time cloud service," \begin{slshape}GRID\end{slshape}, pp.12-21, 2012.



%37
\bibitem{IEEEhowto:X. Wang}
X. Y. Wang, M. Yang, T. Cour and S. H. Zhu, K. Yu, and T. X. Han, ``Contextual weighting for vocabulary tree based image retrieval," \begin{slshape}ICCV\end{slshape}, pp.209-216, 2011.

%38
\bibitem{IEEEhowto:B. Kulis}
B. Kulis and K. Grauman, ``Kernelized locality-sensitive hashing for scalable image search," \begin{slshape}ICCV\end{slshape}, pp.2130-2137, 2009.



%41
\bibitem{IEEEhowto:Z. Zhang}
Z. Zhang, D. S. Katz, J. M. Wozniak, A. Espinosa and I. Foster, ``Design and analysis of data management in scalable parallel scripting," \begin{slshape}SC\end{slshape}, pp.1-12, 2012.



%43
\bibitem{IEEEhowto:M. Almeer}
M. Almeer, ``Cloud Hadoop MapReduce for remote sensing image analysis," \begin{slshape}Journal of Emerging Trends in Computing and Information Sciences\end{slshape}, vol.3, no.4, pp.637-644, 2012.



%47
\bibitem{IEEEhowto:J. Deng}
J. Deng, W. Dong, R. Socher and L.-J.Li \begin{slshape}et al\end{slshape}, ``ImageNet: a large-scale hierarchical image database," \begin{slshape}CVPR\end{slshape}, pp.248-255, 2009.



%49
\bibitem{IEEEhowto:C. Harris}
C. Harris and M. Stephens, ``A combined corner and edge detector," \begin{slshape}Alvey Vision Conference\end{slshape}, 1988.

%新加的参考文献
\bibitem{IEEEhowto:T. L. Liu}
T. L. Liu, D. C. Tao, and D. Xu, ``Dimensionality-Dependent Generalization Bounds for $k$-Dimensional Coding Schemes", Neural Computation (NC), 2016.

\bibitem{IEEEhowto:T. l. Liu}
T. l. Liu, D. C. Tao, M. L. Song, and S. J. Maybank, ``Algorithm-Dependent Generalization Bounds for Multi-Task Learning", IEEE Transactions on Pattern Analysis and Machine Intelligence (T-PAMI), 2016.

\bibitem{IEEEhowto:C. Xu}
C. Xu, D. C. Tao, and C. Xu, ``Multi-view Intact Space Learning", IEEE Transactions on Pattern Analysis and Machine Intelligence , vol. 37, no. 12, pp. 2531-2544, 2015.

\bibitem{IEEEhowto:L. Zhao}
L. Zhao, X. B. Gao, D. C. Tao, X. L. Li, ``Learning a Tracking and Estimation Integrated Graphical Model for Human Pose Tracking", IEEE Transactions on Neural Networks and Learning Systems,  vol. 26, no. 12, pp. 3176-3186, 2015.


\end{thebibliography}
%
% <OR> manually copy in the resultant .bbl file
% set second argument of \begin to the number of references
% (used to reserve space for the reference number labels box)

% biography section
%
% If you have an EPS/PDF photo (graphicx package needed) extra braces are
% needed around the contents of the optional argument to biography to prevent
% the LaTeX parser from getting confused when it sees the complicated
% \includegraphics command within an optional argument. (You could create
% your own custom macro containing the \includegraphics command to make things
% simpler here.)

\begin{IEEEbiography}[{\includegraphics[width=1in,height=1.25in,clip,keepaspectratio]{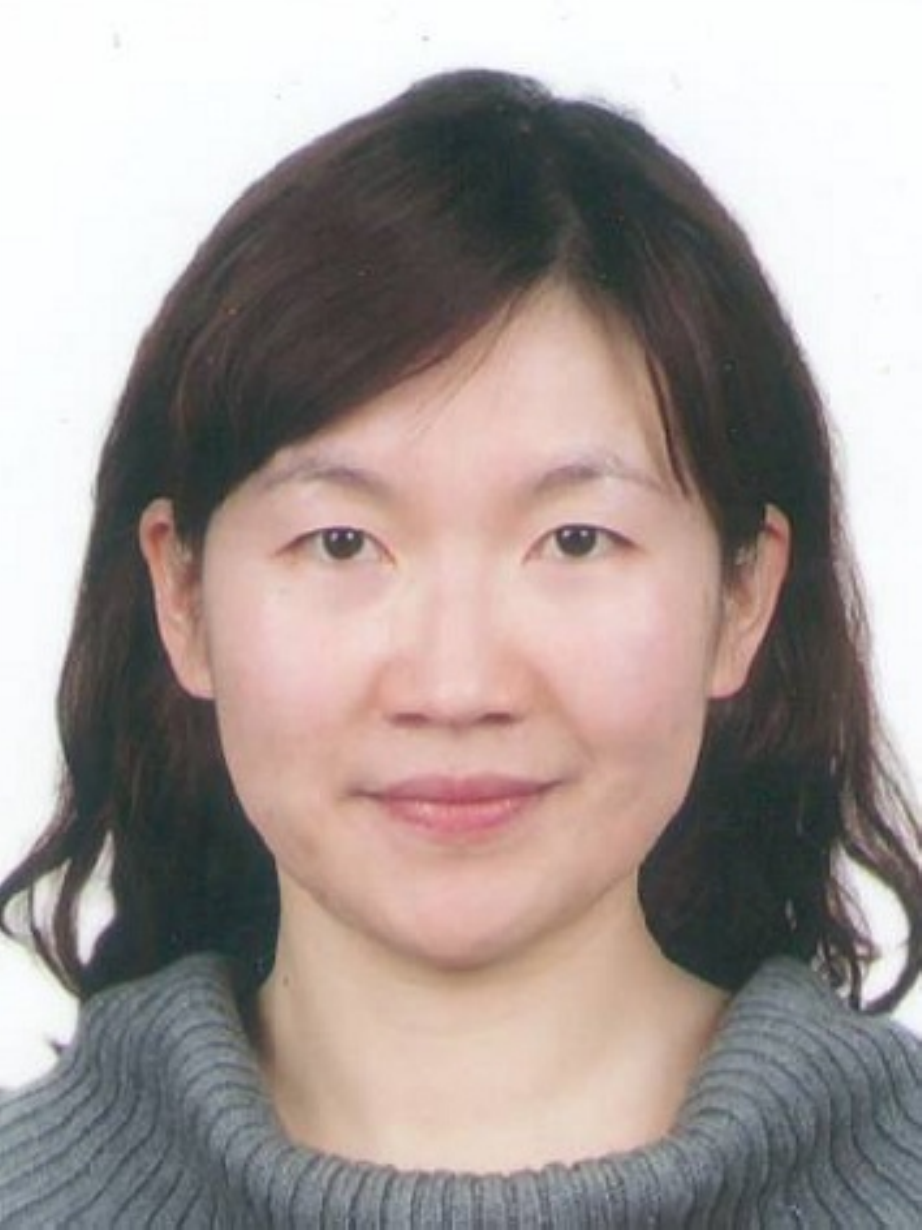}}]{Le Dong}
% or if you just want to reserve a space for a photo:
received the Ph.D. degree from Queen Mary, University of London, London, U.K. She is currently an Associate Professor of Computer Science with the University of Electronic Science and Technology of China, Chengdu, China. She has cochaired/costeered a number of conferences and workshops. Her research areas include bio-inspired models, holons representation framework, mobile social networks, etc., and she is responsible for several projects including NSFC Surface Project, NSFC Youth Project, NSFC Important Research Project, etc.. Till now, Dr. Dong has published papers in reputed international journals and conferences, such as TIP, TCSVT, TNN, TSP, ACM MultiMedia, ICME, ICPR, etc.. Besides, she has served as reviewers for several international journals and conferences including TCSVT, TNN, TIP, PR, etc.. Currently, Dr. Dong is the secretary-general of VALSE, the executive secretary of next-generation National-Local Joint Engineering Center, the director of International Talents Program, HEIFER.
\end{IEEEbiography}

\begin{IEEEbiography}[{\includegraphics[width=1in,height=1.75in,clip,keepaspectratio]{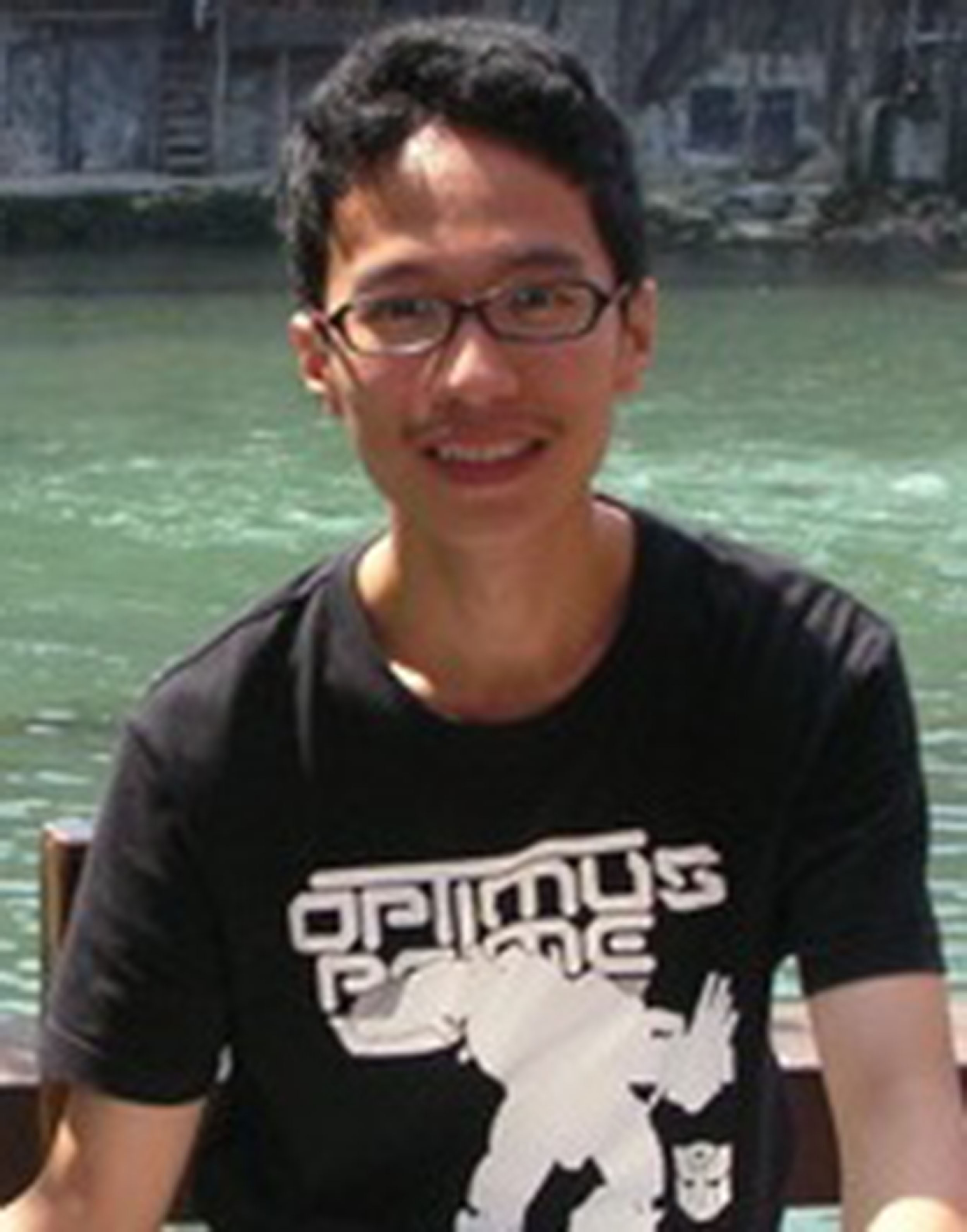}}]{Zhiyu Lin}
% or if you just want to reserve a space for a photo:
received the B.S. degree in computer science and technology and M.S. degree in computer technology from University of Electronic Science and Technology of China, in 2011 and 2014, respectively.
He is working at the Jingdong Chengdu Research Institute.His research interests included image analysis and understanding, big data analysis, machine learning and data mining.
\end{IEEEbiography}

\begin{IEEEbiography}[{\includegraphics[width=1in,height=1.75in,clip,keepaspectratio]{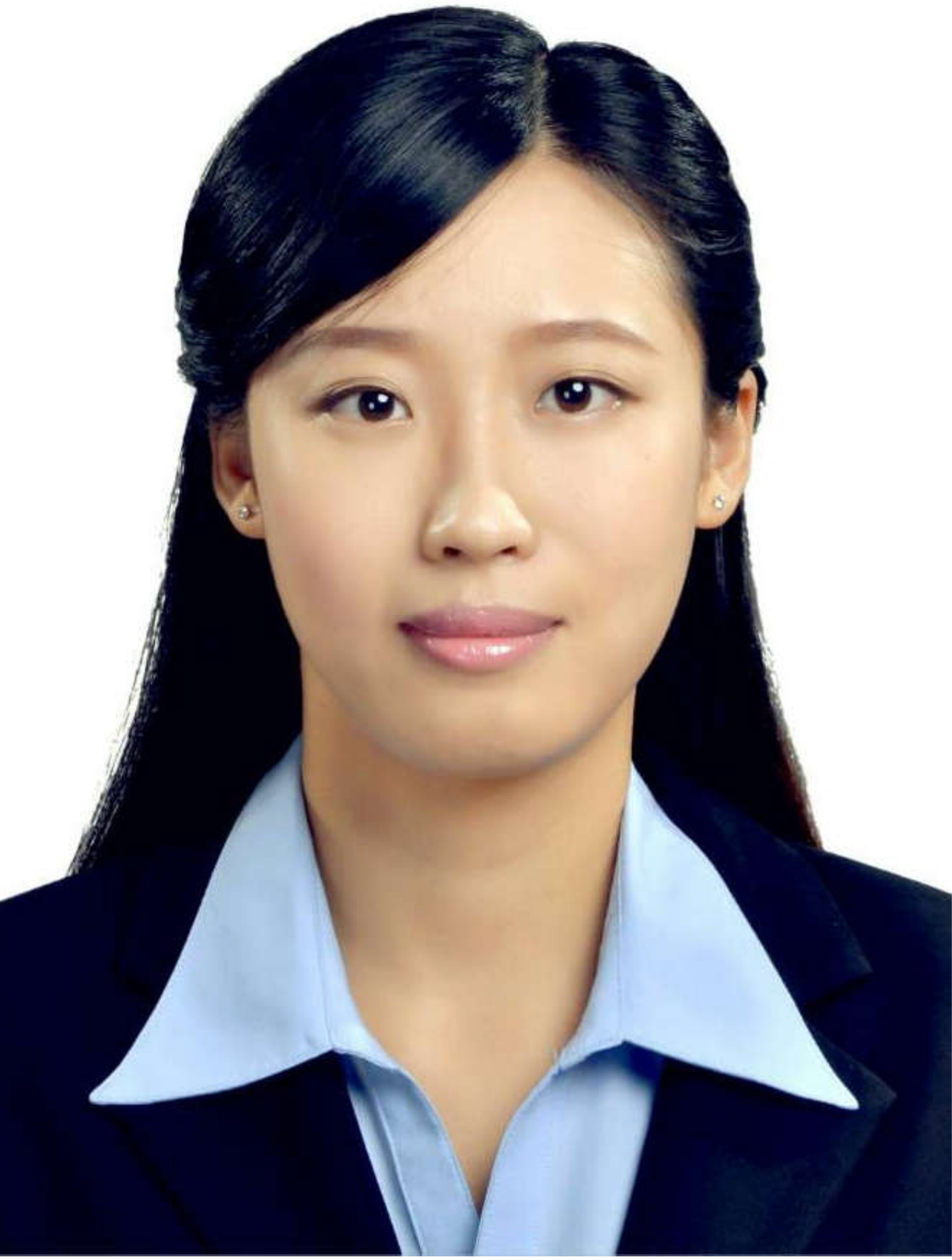}}]{Yan Liang}
% or if you just want to reserve a space for a photo:
received the B.S. degree in computer science and technology and M.S. degree in computer technology from University of Electronic Science and Technology of China, in 2012 and 2015, respectively. She is a Research Staff Member with 123 department in China Academy Of Engineering Physics. Her research interests included image analysis and understanding, big data analysis, machine learning and data mining.
\end{IEEEbiography}

\begin{IEEEbiography}[{\includegraphics[width=1in,height=1.75in,clip,keepaspectratio]{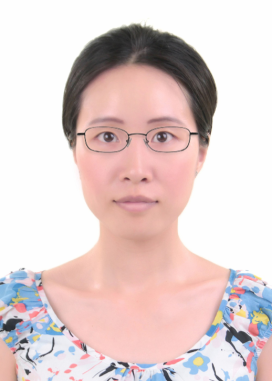}}]{Ling He}
% or if you just want to reserve a space for a photo:
is currently a Postgraduate Student with the School of Computer Science and Engineering,University of Electronic Science and Technology of China, Chengdu, China. Her primary research interests are mainly in image classification, particularly including image clustering, visual matching, and holon representation.
\end{IEEEbiography}
\vfill

\begin{IEEEbiography}[{\includegraphics[width=1in,height=1.75in,clip,keepaspectratio]{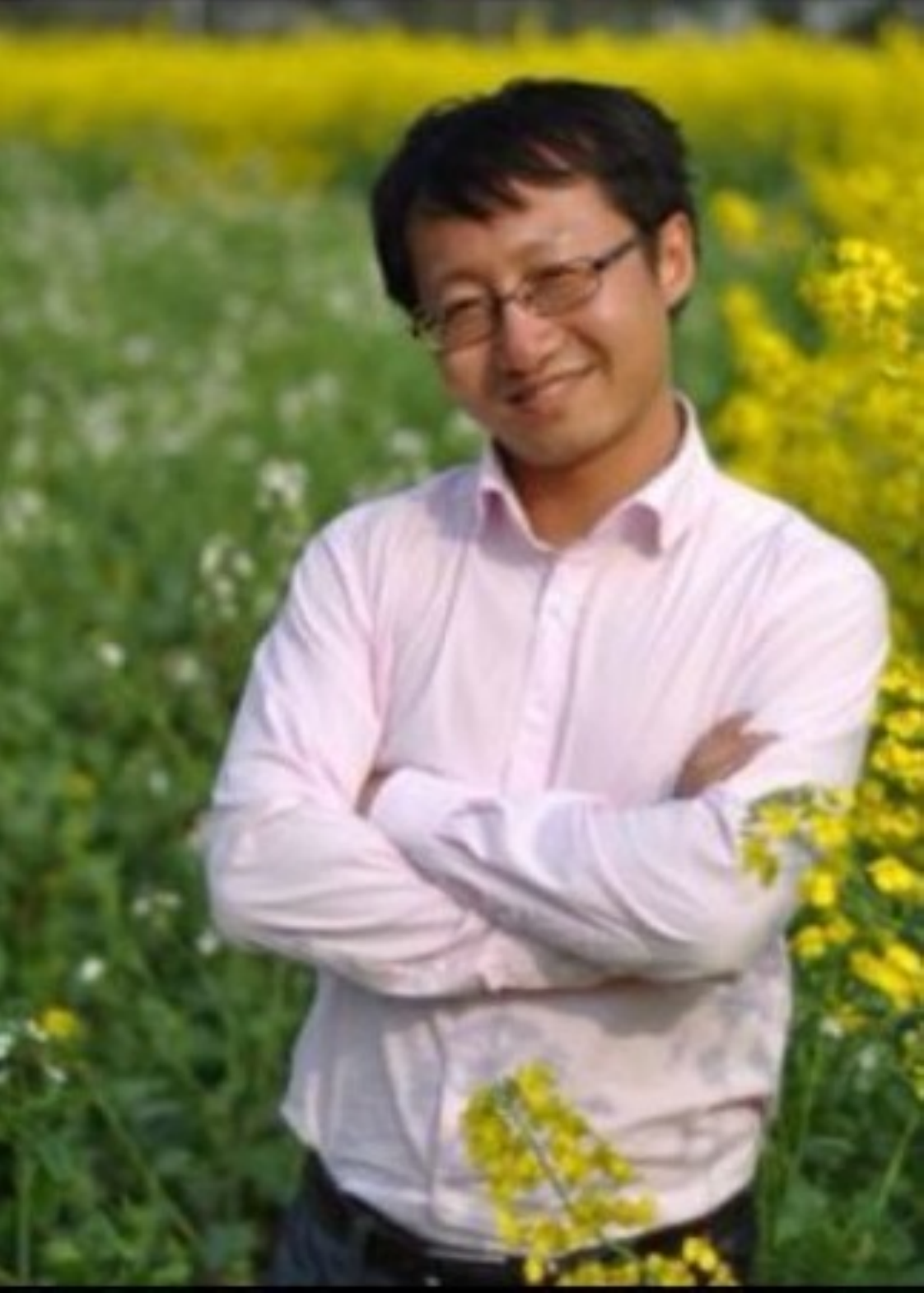}}]{Ning Zhang}
% or if you just want to reserve a space for a photo:
received the Ph.D. degree from University of Electronic Science and Technology of China. He is currently a lecturer of Information and software engineering with the University of Electronic Science and Technology of China, Chengdu, China. His research areas include holons representation framework, embedded systems, O2O framework, etc., and he is responsible for several national projects. Till now, Dr. Zhang has published several papers in related international journals and conferences.

\end{IEEEbiography}

\begin{IEEEbiography}[{\includegraphics[width=1in,height=1.75in,clip,keepaspectratio]{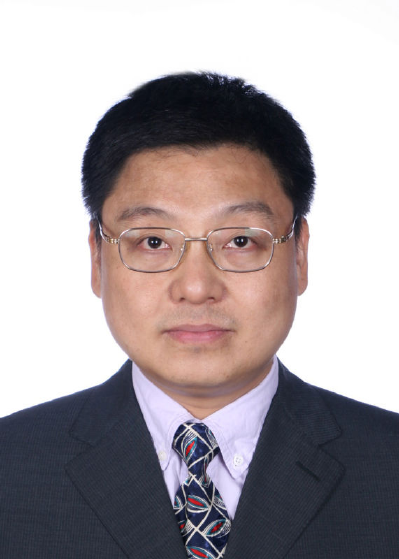}}]{Qi Chen}
% or if you just want to reserve a space for a photo:
is a Senior Engineer with the State Cyberspace Administration of the People's Republic of China. He received the B.S., M.S. and Ph.D. degree in System Engineering from Beijing Jiaotong University. His research interests included network optimization and big data analysis.

\end{IEEEbiography}

\begin{IEEEbiography}[{\includegraphics[width=1in,height=1.75in,clip,keepaspectratio]{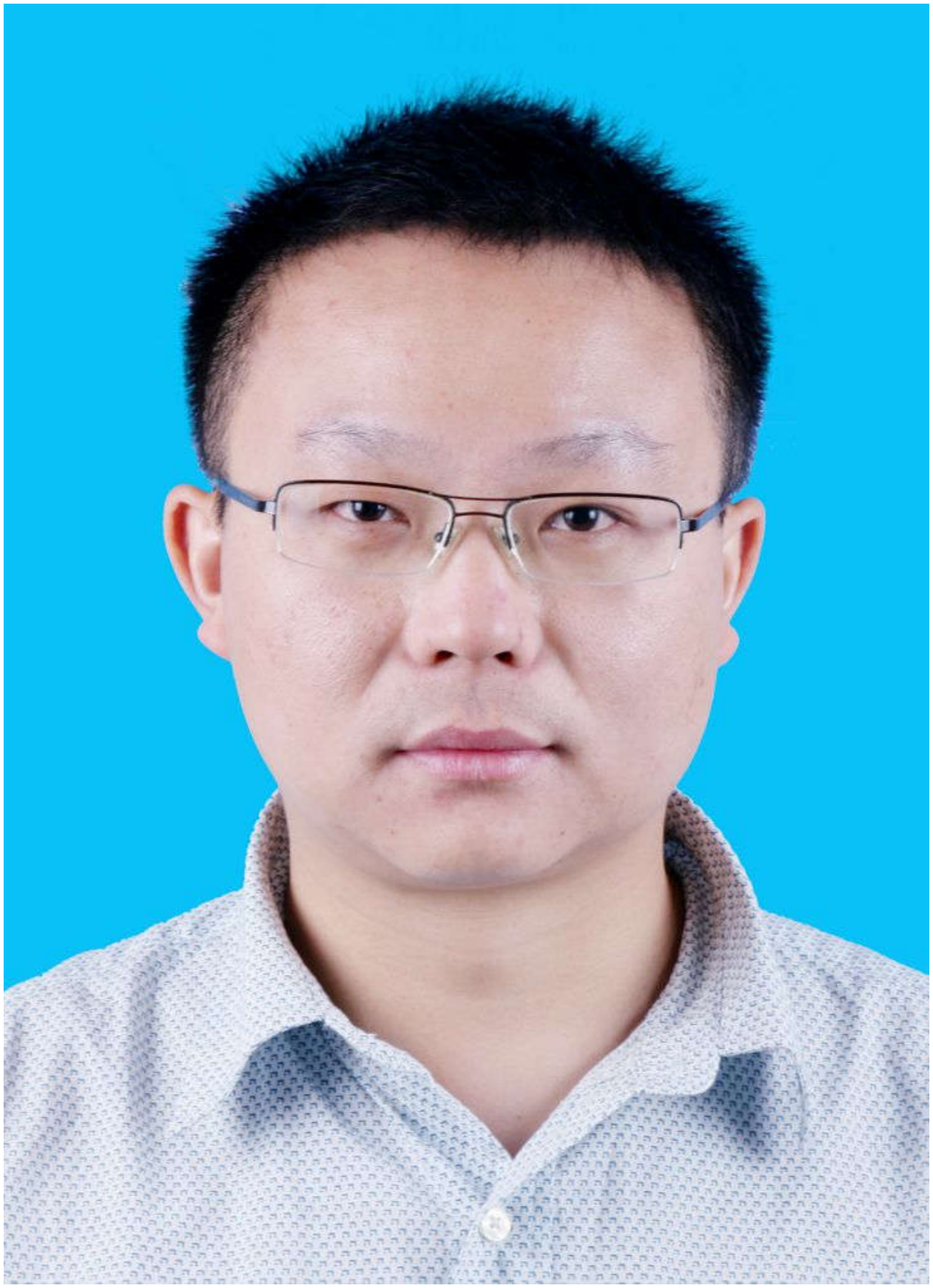}}]{Xiaochun Cao}
% or if you just want to reserve a space for a photo:
 is a professor at the Institute of Information Engineering, Chinese Academy of Sciences since 2012. He received the B.E. and M.E. degrees both in computer science from Beihang University (BUAA), China, and the Ph.D. degree in computer science from the University of Central Florida, USA, with his dissertation nominated for the university level Outstanding Dissertation Award. After graduation, he spent about three years at ObjectVideo Inc. as a
Research Scientist. From 2008 to 2012, he was a professor at Tianjin University. He has authored and coauthored over 80 journal and conference papers. In 2004 and 2010, Dr. Cao was the recipients of the Piero Zamperoni best student paper award at the International Conference on Pattern Recognition. He is on the editorial board of IEEE Transactions of Image Processing. He is a Fellow of IET.
\end{IEEEbiography}

\vfill

\begin{IEEEbiography}[{\includegraphics[width=1in,height=1.75in,clip,keepaspectratio]{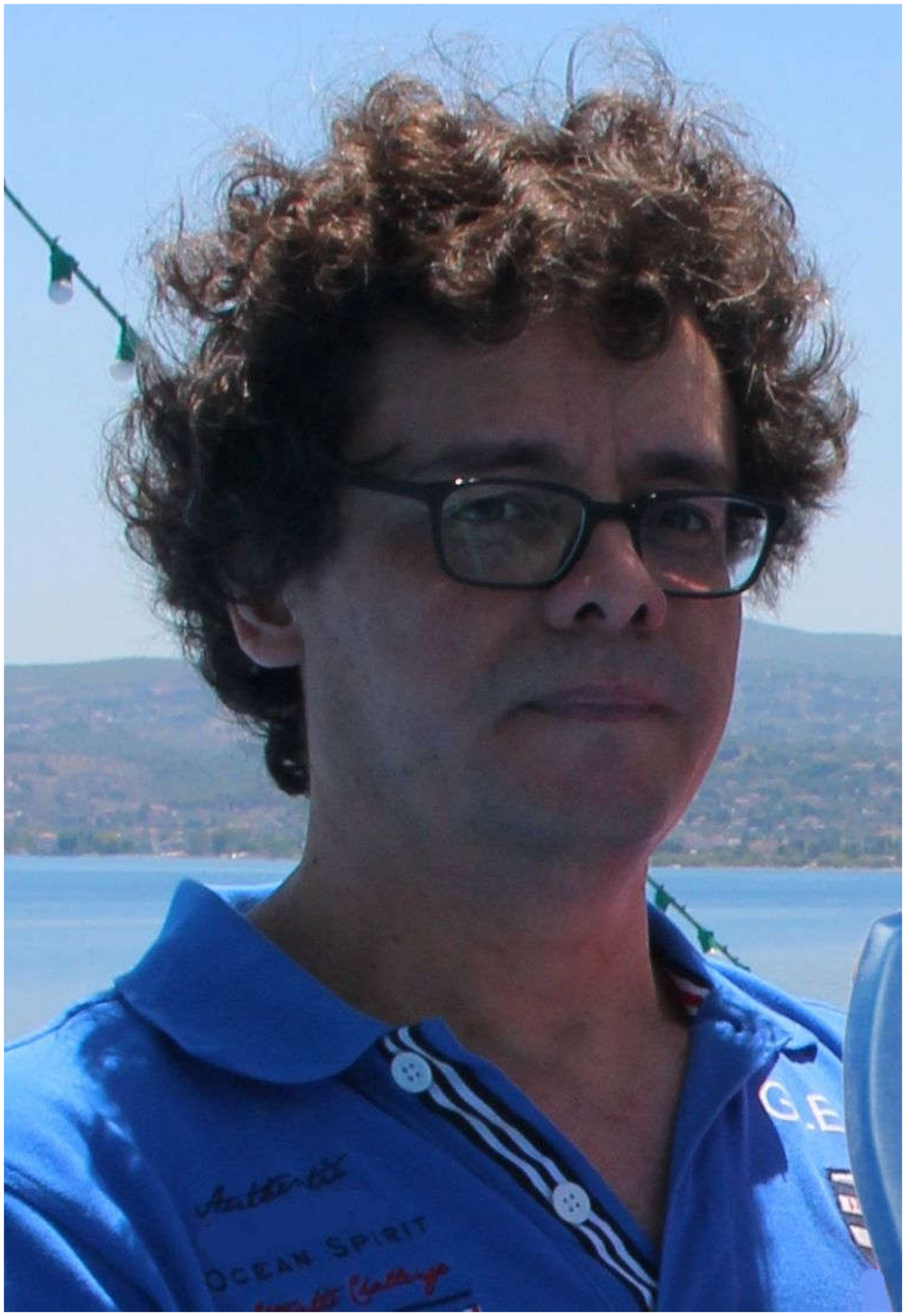}}]{Ebroul Izquierdo}
% or if you just want to reserve a space for a photo:
PhD, MSc, CEng, FIET, SMIEEE, MBMVA, is Chair of Multimedia and Computer Vision and head of the Multimedia and Vision Group in the school of Electronic Engineering and Computer Science at Queen Mary, University of London. He has been a senior researcher at the Heinrich-Hertz Institute for Communication Technology (HHI)), Berlin, Germany, and the Department of Electronic Systems Engineering of the University of Essex. Prof. Izquierdo is a Chartered Engineer, a Fellow member of The Institution of Engineering and Technology (IET), a senior member of the IEEE and a member of the British Machine Vision Association. He was a past chairman of the IET professional network on Information Engineering. He is a member of the Visual Signal Processing and Communications Technical Committee of the IEEE Circuits and Systems Society and member of the Multimedia Signal Processing technical committee of the IEEE.
Prof. Izquierdo is or has been associated editor of the IEEE Transactions on Circuits and Systems for Video Technology (from 2002 to 2010), the IEEE Transactions on Multimedia (from 2010 to 2015). He is member of the editorial board of the
EURASIP Journal on Image and Video processing (from 2004 to date), the Journal of Multimedia Tools and Applications (2008 to 2014) and the Journal of Multimedia (2009-2014), the Journal of Computer Engineering International (2008 to date) and
the Infocommunications Journal (2008 to 1015). He has been guest editor of the Elsevier journal Signal Processing: Image Communication, The EURASIP Journal on Applied Signal Processing and the IEE Proceedings on Vision, Image and Signal
Processing.
\end{IEEEbiography}

\vfill

% if you will not have a photo at all:

% insert where needed to balance the two columns on the last page with
% biographies
%\newpage

% You can push biographies down or up by placing
% a \vfill before or after them. The appropriate
% use of \vfill depends on what kind of text is
% on the last page and whether or not the columns
% are being equalized.

%\vfill

% Can be used to pull up biographies so that the bottom of the last one
% is flush with the other column.
%\enlargethispage{-5in}

% that's all folks
\end{document}